\documentclass{article}

\usepackage{arxiv}
\usepackage{tikz}

\usepackage[utf8]{inputenc} 
\usepackage[T1]{fontenc}    
\usepackage{amsmath,amsfonts}
\usepackage{array}
\usepackage{textcomp}
\usepackage{stfloats}
\usepackage{url}
\usepackage{verbatim}
\usepackage{graphicx}
\usepackage{cite}
\usepackage{color}
\usepackage{amssymb}
\usepackage[figuresright]{rotating}
\usepackage{caption}
\usepackage{subcaption}
\usepackage{multirow}
\usepackage{algpseudocode}

\usepackage[unicode=true]{hyperref}
\hypersetup{
            pdfborder={0 0 0},
            breaklinks=true}
\usepackage{xcolor}
\usepackage{adjustbox}

\usepackage{amsmath,amssymb,amsfonts}
\usepackage{algorithm,algpseudocode}
\usepackage{subcaption}
\usepackage{multirow}

\definecolor{RoyalBlue}{RGB}{65, 105, 225}

\usepackage{pifont}

\usepackage{tabularx}
\usepackage{xcolor}
\definecolor{RoyalBlue}{RGB}{65, 105, 225}

\definecolor{iccvblue}{rgb}{0.21,0.49,0.74}

\usepackage{pifont}
\newcommand{\cmark}{\ding{51}}%
\newcommand{\xmark}{\ding{55}}%

\usepackage{amsmath,amssymb,amsfonts}
\usepackage{algorithm,algpseudocode}
\usepackage{multirow}
\usepackage{subcaption}

\title{Beta Distribution Learning for Reliable Roadway Crash Risk Assessment}

\author{
    Ahmad Elallaf\textsuperscript{\rm 1}\hspace{3mm}
    Nathan Jacobs\textsuperscript{\rm 2}\hspace{3mm}
    Xinyue Ye\textsuperscript{\rm 3}\hspace{3mm}
    Mei Chen\textsuperscript{\rm 4}\hspace{3mm}    
    Gongbo Liang\textsuperscript{\rm 1}\\ [0.75ex]
    \textsuperscript{\rm 1}Texas A\&M University-San Antonio\hspace{3mm}
    \textsuperscript{\rm 2}Washington University in St. Louis\\
    \textsuperscript{\rm 3}University of Alabama \hspace{3mm}
    \textsuperscript{\rm 4}University of Kentucky\\ [0.75ex]
    \texttt{
    \{aelallaf, gliang\}@tamusa.edu\hspace{3mm}
    jacobsn@wustl.edu\hspace{3mm}
    xye10@ua.edu\hspace{3mm}
    mei.chen@uky.edu
    }\\
    [0.75ex]
    \url{https://www.gb-liang.com/projects/betarisk}
	}
\begin{document}
\maketitle
\thispagestyle{firstpage}

\begin{abstract}
Roadway traffic accidents represent a global health crisis, responsible for over a million deaths annually and costing many countries up to 3\% of their GDP. Traditional traffic safety studies often examine risk factors in isolation, overlooking the spatial complexity and contextual interactions inherent in the built environment. Furthermore, conventional Neural Network-based risk estimators typically generate point estimates without conveying model uncertainty, limiting their utility in critical decision-making. To address these shortcomings, we introduce a novel geospatial deep learning framework that leverages satellite imagery as a comprehensive spatial input. This approach enables the model to capture the nuanced spatial patterns and embedded environmental risk factors that contribute to fatal crash risks. Rather than producing a single deterministic output, our model estimates a full Beta probability distribution over fatal crash risk, yielding accurate and uncertainty-aware predictions---a critical feature for trustworthy AI in safety-critical applications. Our model outperforms baselines by achieving a 17-23\% improvement in recall, a key metric for flagging potential dangers, while delivering superior calibration. By providing reliable and interpretable risk assessments from satellite imagery alone, our method enables safer autonomous navigation and offers a highly scalable tool for urban planners and policymakers to enhance roadway safety equitably and cost-effectively.

\end{abstract}  
\section{Introduction}

Roadway traffic accidents claim over 1.3 million lives annually~\cite{who2023road} and impose economic burdens of 3\% of the GDP in many countries~\cite{who2018global}. 
As a critical infrastructure sector~\cite{cisa2024critical}, transportation safety has garnered significant research~\cite{caliendo2007crash,tamerius2016precipitation, zhu2024equity}, yet accurately estimating crash risk remains a challenge due to its inherent uncertainties and the sparse nature of crash events.

Conventional safety research often analyzes individual factors separately, such as driver behavior~\cite{simons2014keep}, road infrastructure~\cite{pembuain2019effect}, traffic patterns~\cite{huang2020highway}, and weather~\cite{jaroszweski2014influence}, overlooking the complex interplay between these elements~\cite{gu2022multivariate}. Since crash occurrences frequently result from intricate multi-factor interactions, methods that analyze these factors in isolation struggle to predict risk holistically~\cite{carrodano2024data}. Furthermore, data limitations have constrained the scope of most studies to highways~\cite{ahmed2013road, song2018farsa, cheng2019risk, ma2020modeling, joo2023a}, leaving comprehensive crash risk analysis for local roads, where data is often less available, relatively unexplored.

To overcome these limitations, we introduce a novel deep learning framework that learns a full Beta probability distribution, moving beyond simple point-estimates of fatal crash risk. Our primary contributions are threefold:
\begin{itemize}
    \item A \textbf{holistic, vision-based model} that captures the complex interplay of risk factors embedded in the visual data, in contrast to methods that study variables in isolation.

    \item A \textbf{probabilistic formulation} that yields well-calibrated, uncertainty-aware predictions, a critical feature for trustworthy AI in high-stakes, safety-critical domains.

    \item A \textbf{highly scalable and equitable methodology} that uses near-globally available satellite imagery, enabling risk assessment for both highways and previously under-assessed local roads.
\end{itemize}

The proposed probabilistic model is evaluated through extensive experiments conducted over four major metropolitan areas, which have a population of $\approx20$ million. Our model achieves a 17-23\% improvement in recall over baselines, a crucial metric for any safety-critical task, while also delivering superior model calibration and F1 scores. By producing reliable and interpretable risk assessments from satellite imagery alone, this work provides a foundational tool for enhancing traffic safety, from enabling safer route selection for drivers and autonomous vehicles to empowering urban planners and policymakers to mitigate high-risk areas.

\section{Background}

\subsection{Estimate Roadway Crash Risk} 
\label{sec:task_challenges}

A primary challenge in data-driven roadway safety is formulating the risk estimation task. Existing methods often frame it as classifications, such as predicting a crash occurrence within a short time frame~\cite{huang2020highway}. While valuable, these approaches do not estimate the inherent, continuous crash risk of a given road segment. A more nuanced approach is to directly estimate a crash probability, such as using Monte Carlo simulations~\cite{de2018evaluating,al2012use,jeon2016monte}. However, this is fundamentally challenged by the extreme sparsity of crash data. For instance, the average annual accident rate for a 25m$^2$ road segment in the United States is just $0.1\%$~\cite{moosavi2019accident}. This level of sparsity renders traditional estimation techniques unreliable, as they can obscure high-risk areas while falsely flagging safe ones~\cite{he2021inferring}, leading to false negatives that are dangerous in any safety-critical application. Furthermore, such simulation methods are often ill-suited for large-scale applications due to high computational costs and the need for carefully tuned parameters.

Deep Neural Networks (DNNs) offer a powerful alternative, as they can learn complex, task-specific features directly from data and provide near-instantaneous inference. However, supervised DNNs typically rely on large, manually labeled datasets, such as manually assigned risk levels (e.g., low, neutral, high)~\cite{najjar2017combining}. Creating these datasets is prohibitively expensive, and the manual labels can suffer from human bias, potentially misrepresenting the true risk~\cite{li2024label,chen2023ai}. These challenges motivate the need for a new approach that can learn a continuous risk score from objective crash data while effectively handling the probabilistic nature of the task.

\subsection{Deep Neural Network Miscalibration}

Over the recent years, DNNs have shown promising performance on various domains, such as medical imaging~\cite{xing2023self,liu2022llrhnet}, cybersecurity~\cite{zulu2024enhancing}, transportation~\cite{jonnala2025exploring}, and astrophysics~\cite{lin2022estimating}. However, for a predictive model to be trustworthy in high-stakes applications, its predicted confidence must accurately reflect its probability of being correct. However, modern DNNs are often miscalibrated, tending to produce overconfident predictions~\cite{pereyra2017regularizing, guo2017calibration}. 

Mathematically, a model is perfectly calibrated if, for any given confidence level $p$, the long-run accuracy of predictions with that confidence is indeed $p$. For DNNs, the calibration error, the difference between a model's predicted confidence and its actual accuracy, is often significantly greater than zero~\cite{hinton2015distilling}. This miscalibration is a critical failure point in high-stakes applications where decisions depend on the model's self-assessed certainty.

While various techniques can mitigate this issue, they often have limitations. Post-processing methods like temperature scaling~\cite{guo2017calibration} adjust model outputs without altering the learned features, while in-training regularization~\cite{kumar2018trainable,liang2020imporved} requires careful tuning for the weight scaler. Given that model complexity is a key contributor to miscalibration~\cite{chidambaram2023on}, we argue that an effective solution must be deeply integrated into the learning process. Our work achieves this by reformulating the risk estimation task as learning a full probability distribution, a method that inherently encourages better-calibrated and more reliable predictions.
\section{Method}

\subsection{Probabilistic Modeling Framework} 

Our method recasts roadway crash risk estimation from a standard classification task into a probabilistic learning problem, motivated by the limitations of conventional models that provide a single point-estimation.
Consider a fatal crash, a stochastic occurrence, at a specific point in spacetime, $C = (x, y, t, d)$, where $(x,y)$ is the geolocation and $(t,d)$ is the time and date. While any single crash is a random event, its location provides the strongest available evidence for a local maximum in the underlying, continuous risk field, $R(\cdot)$. Therefore, it is intuitive that the inferred risk should be higher at or near the crash site and should decay smoothly as one moves away in space or time. For nearby points, such as a spatially displaced point $C' = (x-\delta, y, t, d)$, 
the risk should be lower, i.e., $R(C') < R(C)$. 
Standard point-estimate classifiers fail to capture this continuous field, as they are trained to predict a binary outcome for each location independently.

While a complete model would account for both spatial and temporal decay, this work focuses on the challenging and foundational task of estimating the \textbf{static, inherent risk} of a location based on its geographic and structural features. 
Our goal is to model the spatial component of this uncertainty by learning a distribution over possible risk values, capturing the intuition that for a nearby point $C'$, the risk is attenuated but non-zero: $0<R(C')<R(C)$.

We specifically employ the Beta distribution for this task due to its natural support on the $[0, 1]$ interval and its flexibility in representing diverse risk profiles. Instead of a single value, our model $h(x)$ maps an input image $x$ to the two positive scalar parameters, $(\alpha, \beta)$, which define a Beta distribution, $P_p\sim Beta(\alpha, \beta)$. This formulation allows the model to express its uncertainty through the shape of the distribution: a sharp peak indicates high confidence, while a wide distribution signifies high uncertainty. 
The final risk score $R$ is the mean of this predicted distribution:
\begin{equation}    
\label{eq:risk}
    R = \mathbb{E}[P_p] = \frac{\alpha}{\alpha + \beta}. 
\end{equation}

To achieve this, our framework integrates three key technical contributions: 1) a novel procedural labeling technique that generates the targeting Beta distributions from data augmentation, 2) a multi-scale deep neural network architecture, and 3) a compound loss function for joint optimization.

\subsection{Target Beta Distributions Generation}

A key innovation of our framework is the procedural generation of supervisory signals in the form of target Beta distributions. Instead of using static labels, we dynamically create a target Beta distribution, $P_t\sim Beta(\alpha_t, \beta_t)$, for each training sample based on the properties of the random crop augmentation. Specifically, given an input image, we first apply a random crop. The target Beta distribution is, then, generated using Algorithm~\ref{alg:beta_label}. This process acts as a sophisticated form of structured label smoothing, transforming data augmentation from a simple regularizer into a rich source of continuous supervision for risk and uncertainty.

\begin{algorithm}[!tb]
    \caption{Target Beta Distribution Generation}
    \begin{algorithmic}

    \Require Original image $x$, binary label $l \in \{0, 1\}$, base concentration $K_{base}$, minimum positive risk mean $\mu_{min}$, minimum positive concentration $k_{min}$, distance weight $w_{dist}$, size weight $w_{size}$, and $\epsilon=1e^{-5}$

    \vspace{1mm}
    \If{$l = 0$}  \Comment{For negative samples, create a low-risk,}
    \State \Comment{high-certainty distribution}
    
        \State $\alpha_t \gets \epsilon$ 
        \State $\beta_b \gets K_{base}$ 

    \vspace{.5mm}
    \Else \Comment{For positive samples (l=1), generate labels}
    \State \Comment{based on crop geometry}
        \State $x' \gets$ random crop of $x$
        
        \State $d_{norm} \gets$ normalized distance of $x'$ from center of $x$
        
        \State $s_{norm} \gets \frac{size(x')}{size(x)}$
        
        \State $influence \gets w_{dist} \cdot (1 - d_{norm}) + w_{size} \cdot s_{norm}$
        
        \State $\mu_t \gets \mu_{min} + (1 - \mu_{min}) \cdot influence$
        
        \State $k_t \gets k_{min} + (K_{base} - k_{min}) \cdot influence$
        
        \State $\alpha_t \gets \mu_t \cdot k_t$
        
        \State $\beta_t \gets \epsilon$

    \vspace{.5mm}
    \EndIf \\

    \vspace{.5mm}
    \Return $(\alpha_t, \beta_t)$ \Comment{Return the target Beta distribution}    
    
    \end{algorithmic}
    \label{alg:beta_label}    
\end{algorithm}

For \textbf{negative samples} (no crash), the objective is to predict low risk with high confidence. The target distribution is therefore constant: $\alpha_t$ is set to a small positive value $\epsilon$ and $\beta_t$ is set to a large value representing high certainty $K_{base}$, creating a distribution sharply peaked at zero.

For \textbf{positive samples} (crash), the target distribution reflects the quality of the visual evidence in the random crop. This is quantified by an \texttt{influence} score, which modulates the target distribution's mean and concentration to generate a supervision signal that is proportional to the information content of the augmented image. The score is a weighted combination of two geometric properties of the crop: its centrality relative to the crash location and its size.

We set the weights to 0.7 for centrality ($w_{dist}$) and 0.3 for relative size ($w_{size}$). This weighting scheme is based on the strong intuition that the visual features most critical to understanding risk--such as specific road geometry, lane markings, the presence of an intersection, or the surrounding environments--are spatially concentrated around the event's location. A crop that is well-centered on the crash point provides the clearest and most relevant evidence, thus deserving a higher \texttt{influence} score and a more confident target distribution. The relative size of the crop provides useful, but secondary, broader context about the surrounding environment. This principled approach transforms data augmentation into a rich source of supervision, teaching the model to dynamically associate higher risk and confidence with visual samples that contain the most informative evidence.

This \texttt{influence} score then modulates the target mean $\mu_{min}$ and the target concentration $k_t$, which in turn define the final Beta parameters. For positive samples, the $\beta_t$ is set to the small constant $\epsilon$, ensuring the distribution is always skewed towards high risk, with the \texttt{influence} score controlling the precise shape and confidence (see the \textbf{supplement materials} for the hyperparameters used in this study).

\subsection{Model Architecture}

\begin{figure}[!tb]
    \centering
    \includegraphics[width=0.9\textwidth]{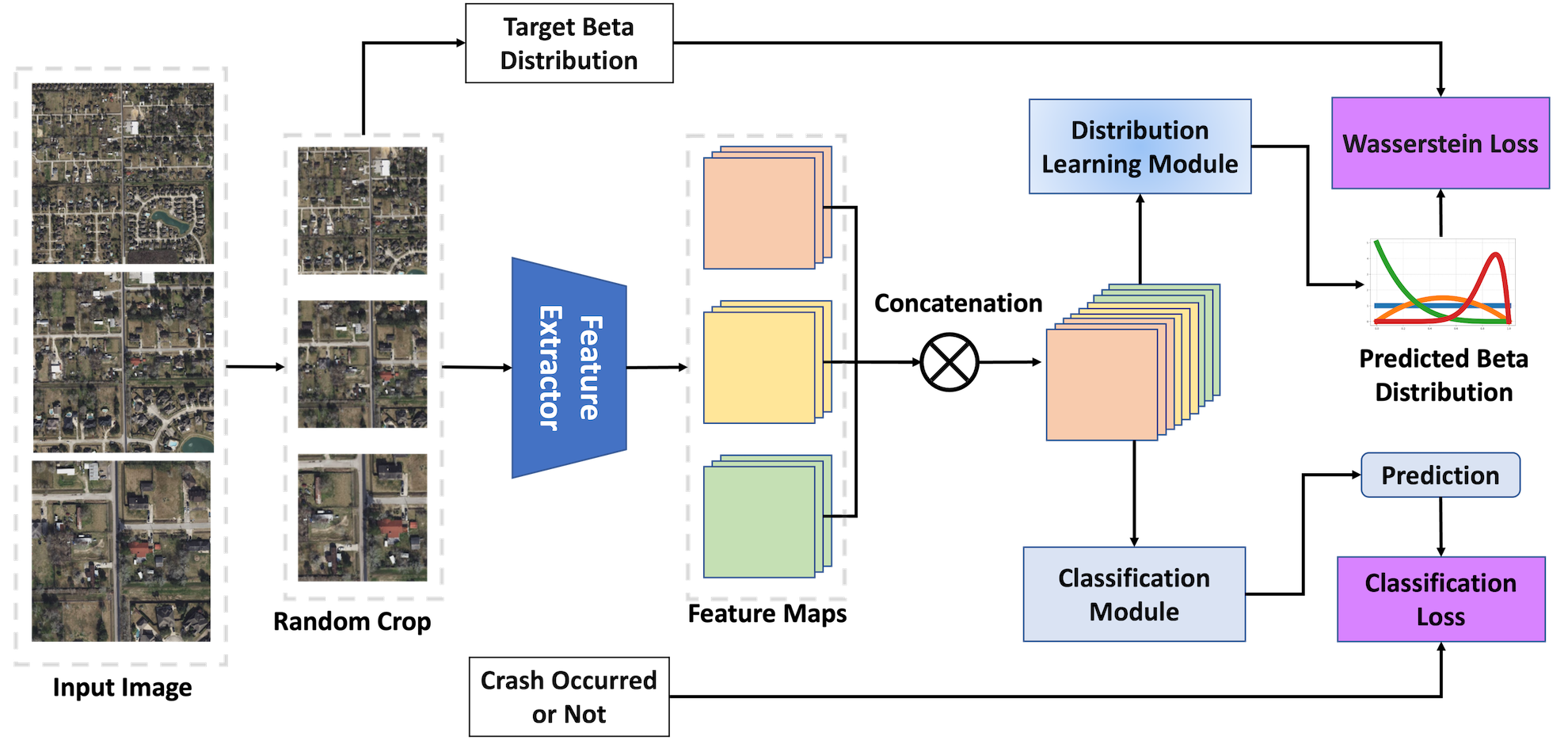} 
    \caption{Training Architecture with Joint Optimization}
    \label{fig:architecture}
\end{figure}

The architecture of our model, illustrated in Figure~\ref{fig:architecture}, is designed to process multi-scale satellite imagery. During training, a random crop is sampled from the input, which consists of image slices of the same location at different resolutions.

The cropped images are, then, passed through a shared feature extractor backbone to produce multiple corresponding feature maps. These maps are concatenated along the channel dimension to form a unified feature representation, serving as the input for two parallel prediction heads:
\begin{itemize}
    \item A Distribution Learning Head, which outputs the two Beta parameters $(\alpha, \beta)$.
    \item An auxiliary Classification Head, which outputs a single logit for the binary crash/no-crash task.
\end{itemize}

\subsection{Training and Optimization}
The model is trained end-to-end by jointly optimizing the two parallel heads with a compound loss function. 
The primary distribution learning head is supervised by the a mean-variance loss that inspired by the squared Wasserstein-2 ($W^2_2$) distance~\cite{vaserstein1969markov}, which measures the dissimilarity between the predicted ($P_p$) and the target ($P_t$) Beta distributions:
\begin{equation}
    \mathcal{L}_{W_2^2}(P_p, P_t) = (\mu_p - \mu_t)^2 + (\sigma_p - \sigma_t)^2,
\end{equation}
where the $\mu$ and $\sigma$ are the mean and standard deviation.

We empirically selected this $W^2_2$ surrogate over true $W^2_2$ distance and other distribution divergence metrics, including KL-Divergence~\cite{csiszar1975divergence} and the Cramér-von Mises criterion~\cite{cramer1928on}. As a true metric, our $W^2_2$ surrogate loss provides a more stable gradient than KL-Divergence, especially when the predicted and target distributions have little overlap. Most importantly, for one-dimensional distributions like the Beta, the $W^2_2$ surrogate loss directly optimize of the risk score (the mean) and confidence level (the standard deviation) simultaneously. Our experimental analysis also shows this surrogate is a close approximation of the true $W^2_2$ (errors on the order of $10^{-3}$ to $10^{-2}$), deviating only in extreme cases (see the \textbf{supplementary materials}).

The auxiliary classification head is supervised by a Binary Cross-Entropy loss, which encourages the shared backbone to learn discriminative features relevant to the safety task:
\begin{equation}
    \mathcal{L}_{BCE} = -\frac{1}{N} \sum_{i=1}^{N} \left[ y_i \log(p_i) + (1-y_i) \log(1-p_i) \right],   
\end{equation}
where $y_i$ and $p_i$ are the label and predicted probability.

The overall objective function is a weighted combination of the two losses, balanced by hyperparameters, $\lambda_1$ and $\lambda_2$:
\begin{equation}
	\mathcal{L} = \lambda_1 \cdot \mathcal{L}_{BCE} + \lambda_2 \cdot \mathcal{L}_{W_2^2}.
\end{equation}

\subsection{Inference Process}
\begin{figure}[!tb]
    \centering
    \includegraphics[width=0.9\textwidth]{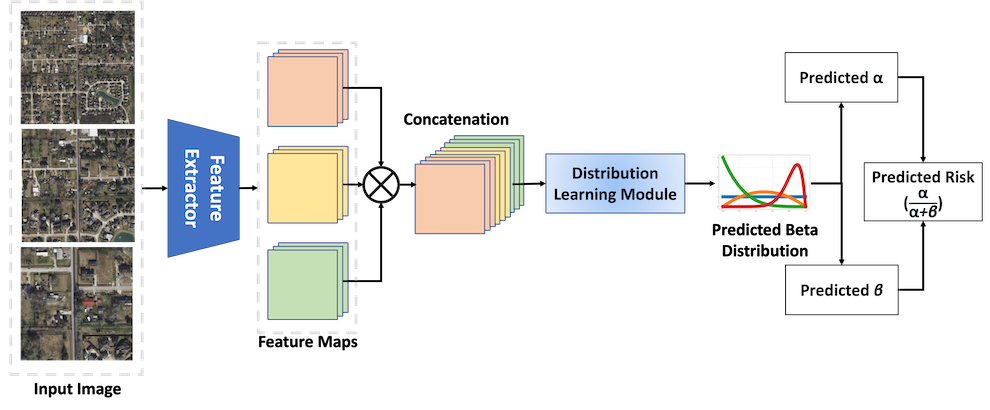}  
    \caption{Streamlined Architecture for Inference}
    \label{fig:architecture_inference}
\end{figure}

The inference process, illustrated in Figure~\ref{fig:architecture_inference}, is direct and computationally efficient. The random crop augmentation and the auxiliary classification head are removed. The full, uncropped multi-scale image is passed through the feature extractor backbone and the distribution head. The risk score $R$ is calculated as the mean of the distribution, per Equation~\ref{eq:risk}.
This feed-forward process allows for rapid and scalable risk assessment of any location.

\section{Experiment Setup}

This study utilizes the MSCM dataset~\cite{liang2024unveiling}, a large-scale collection of multi-scale satellite images from Texas, USA, with 16,451 locations labeled with historical fatal crashes. All models use a ResNet-50~\cite{he2016deep} backbone, $\lambda_1=5, \lambda_2=1$, and were trained on NVIDIA A100 GPUs. See the \textbf{supplementary materials} for more information about the dataset, implementation details, and hyperparameter analysis for the selection of $\lambda_1$ and $\lambda_2$.

\subsection{Evaluation Methodology}

\paragraph{Quantitative Metrics} We first evaluate our model's practical effectiveness by framing the risk estimation as a binary classification task to identify historical crash locations. 
The model's predicted risk score $R$, derived from Equation~\ref{eq:risk} is thresholded at $0.5$ to yield a binary prediction. We then assess the model's predictive performance using standard metrics: F1-Score, Precision, Recall, AUC (Area Under the Receiver Operating Characteristic curve), and PRC (area under the precision-recall curve); and assess model's calibration using Expected Calibration Error (ECE) and Brier score. Due to safety-oriented, \textbf{we consider Recall to be the most critical metric} that answers the question: ``\textit{Of all crash locations, what fraction did our model successfully identify?}"

We also evaluate our method against a Deep Ensemble (DE) of the strongest baseline, constructed from three independent training runs. The final predicted risk score of a DE model, $R_{DE}$, is calculated as the mean of the predictions from each individual model in the ensemble. This single score for each sample is then used to compute all the aforementioned performance and calibration metrics. 

The ensemble's predictive uncertainty is quantified in two ways: the variance of the risk scores and the disagreement rate among the final binary predictions. A higher value in either metric reflects greater disagreement among the models and thus higher uncertainty in the final prediction.

\paragraph{Qualitative Analysis} To intuitively understand the value of our probabilistic approach, we conduct a qualitative analysis of the model's outputs from two perspectives. 
First, we analyze the aggregate behavior of the model's outputs by comparing the overall distribution of predicted probabilities from our model against the baselines. By plotting a histogram of all risk scores, 
we can visually assess model confidence. A well-calibrated, uncertainty-aware model is expected to utilize the full [0, 1] probability range, whereas overconfident models will show predictions heavily clustered at the extremes (near 0 and 1). 
Second, we visualize the predicted Beta distributions for four distinct scenarios: true positives (TP), true negatives (TN), false negatives (FN), and false positives (FP). The goal of this analysis is to provide an intuitive understanding of the model's behavior by interpreting its successes and failures. 

\paragraph{Case Study: San Antonio River Walk} To demonstrate the model's utility, we conduct a case study of the San Antonio River Walk, providing practical insight into the model's performance in a challenging, safety-critical area.



    
    


        


\subsection{Baseline Models}
We evaluate our method against three baselines to isolate our framework's contributions. Our primary benchmark is the Multi-Scale Cross-Matching (MSCM) model~\cite{liang2024unveiling}, the current state-of-the-art for fatal crash risk estimation using only satellite imagery.

\paragraph{\texttt{ImageNet} Baseline:} A standard model pre-trained on ImageNet~\cite{krizhevsky2012imagenet} that takes single-scale satellite images as input, providing us the performance of a generic, non-domain-specific feature extractor on our task.

\paragraph{\texttt{MSCM-SS} (Single-Scale):} The same single-scale architecture but using weights generated by the self-supervised pre-training through cross-matching, proposed in the MSCM paper, to test the value of domain-specific features.

\paragraph{\texttt{MSCM-MS} (Multi-Scale):} The full MSCM model, which uses both its domain-specific pre-training and multi-scale imagery as input, represents the strongest available baseline, allowing us to compare against the current state-of-the-art classification approach directly.

\vspace{1.25mm}
\noindent The best checkpoint for each model was selected based on the model accuracy on the validation set.

\section{Results}

\subsection{Quantitative Analysis}
Table~\ref{table:result} summarizes the quantitative results. The single-scale baselines (\texttt{ImageNet} and \texttt{MSCM-SS}) achieve a $<0.5$ precision and recall scores, indicating their predictions for positive cases are close to random and exhibit little ability to identify high-risk areas. While the \texttt{MSCM-MS} model achieves high precision (0.6731), its poor recall (0.4521) means it fails to identify over half of all crash locations, rendering it unreliable for safety-critical applications.

\begin{table*}[!tb]
    \setlength{\tabcolsep}{2.5pt}
    \small
	\centering
	\caption{Main Quantitative Results ({\bf{bold}}: best performance; {\underline{underlined}}: second best performance)}

    \begin{tabular}{c||c|c|c||c|c|c|c|c||c|c} \hline \hline
    \multirow{2}{*}{\textbf{Methods}} & \multirow{2}{*}{\textbf{Pre-Train}} & \multirow{2}{*}{\textbf{Probabilistic}} & \multirow{2}{*}{\textbf{Multi-Scale}} & \multicolumn{5}{c||}{\textbf{Performance ($\uparrow$)}}  & \multicolumn{2}{c}{\textbf{Uncertainty} ($\downarrow$)}\\\cline{5-11}

    & & & & \textbf{F1} & \textbf{Precision} & \textbf{Recall} & \textbf{AUC}  & \textbf{PRC} & \textbf{ECE} & \textbf{Brier}\\\hline\hline

    \textbf{ImageNet} & ImageNet & \xmark &\xmark & $0.4753$ & $0.4968$ & $0.4555$ & $0.7980$ & $0.4862$ & $0.1281$ & $0.1600$\\\hline
    
    \textbf{MSCM-SS} & MSCM & \xmark & \xmark & $0.4966$ & $0.4981$ & $0.4950$ & $0.8165$ & $0.5185$ & $\underline{0.1006}$ & $0.1458$ \\\hline
    
    \textbf{MSCM-MS} & MSCM & \xmark & \cmark & $\underline{0.5409}$ & $\bf{0.6731}$ & $0.4521$ & $\underline{0.8572}$ & $\underline{0.6269}$ & $0.1067$ & $ \underline{0.1296}$\\\hline

    \textbf{Prob-SS (Ours)} & MSCM & \cmark & \xmark & $0.5001$ & $0.4252$ & $\bf{0.6070}$ & $0.7749$ & $0.4409$ & $0.1731$ & $0.1922$ \\\hline

    \textbf{Prob-MS (Ours)} & MSCM & \cmark & \cmark & $\bf{0.5762}$ & $\underline{0.6296}$ & $\underline{0.5311}$ & $\bf{0.8663}$ & $\bf{0.6489}$ & $\bf{0.0881}$ & $\bf{0.1211}$\\\hline \hline
        
	\end{tabular}
	\label{table:result}
\end{table*}

\begin{table*}[!tb]
    \setlength{\tabcolsep}{3.2pt}
	\centering
    \small

	\caption{Deep Ensemble Results over Three Training Trails ({\bf{bold}}: best performance)}

    \begin{tabular}{c||c|c|c|c|c||c|c||c|c} \hline \hline
    \multirow{2}{*}{\textbf{Methods}} & \multicolumn{5}{c||}{\textbf{Performance ($\uparrow$)}}  & \multicolumn{2}{c||}{\textbf{Uncertainty} ($\downarrow$)} & \multicolumn{2}{c}{\textbf{Disagreement} ($\downarrow$)}\\\cline{2-10}

    & \textbf{F1} & \textbf{Precision} & \textbf{Recall} & \textbf{AUC}  & \textbf{PRC} & \textbf{ECE} & \textbf{Brier} & \textbf{Variance} & \textbf{Disagr. Rate}\\\hline\hline

    \textbf{Ensemble MSCM-MS} & $0.5966$ & $\bf{0.7062}$ & $0.5165$ & $\bf{0.8839}$ & $\bf{0.6890}$ & $0.0787$ & $0.1112$ & $0.0925$ & $16.93\%$\\\hline

    \textbf{Ensemble Prob-MS (Ours)} & $\bf{0.5976}$ & $0.6750$ & $\bf{0.5361}$ & $0.8761$ & $0.6886$ & $\bf{0.0605}$ & $\bf{0.1075}$ & $\bf{0.0822}$ & $\bf{15.14\%}$\\\hline \hline
        
	\end{tabular}
	\label{table:result_ensemble}
\end{table*}

In contrast, our models demonstrate a significant improvement in identifying potential dangers. Our multi-scale model, \texttt{Prob-MS}, achieves the best overall balance of performance, attaining the highest F1-score. Its most significant contribution is boosting the recall to 0.5311, a 17\% relative improvement over \texttt{MSCM-MS}, drastically reducing the number of hazardous sites that would be missed. Our single-scale model, \texttt{Prob-SS} (0.6070 recall score), significantly improves the metric by 23\% over the best baseline (0.4950).

Crucially, \texttt{Prob-MS} is also the most trustworthy model, achieving the lowest (best) ECE of 0.881 and Brier of 0.1211. This confirms that our model's probabilistic outputs are more statistically sound and reliable for real-world decision-making.

We also evaluate our method against a Deep Ensemble of the strongest baseline (Table~\ref{table:result_ensemble}). When comparing our single Prob-MS model against the baseline Ensemble MSCM-MS, we find that our single model achieves competitive performance, including a 3\% higher recall, better calibration, and lower uncertainty at only 1/3 the computational cost at both training and inference times. This highlights the efficiency and practical advantage of our approach.

In an apples-to-apples comparison between ensembled methods, our Ensemble Prob-MS demonstrates the clear superiority of our probabilistic framework. It outperforms the baseline ensemble on the most critical metrics for this task, achieving a higher F1-score and recall. Most importantly, it is significantly better calibrated and exhibits lower uncertainty, as evidenced by its superior (lower) ECE, Brier, Variance, and Disagreement Rate scores.

\subsection{Qualitative Analysis}

\begin{figure*}[!tb]
    \centering
    \includegraphics[width=.95\linewidth]{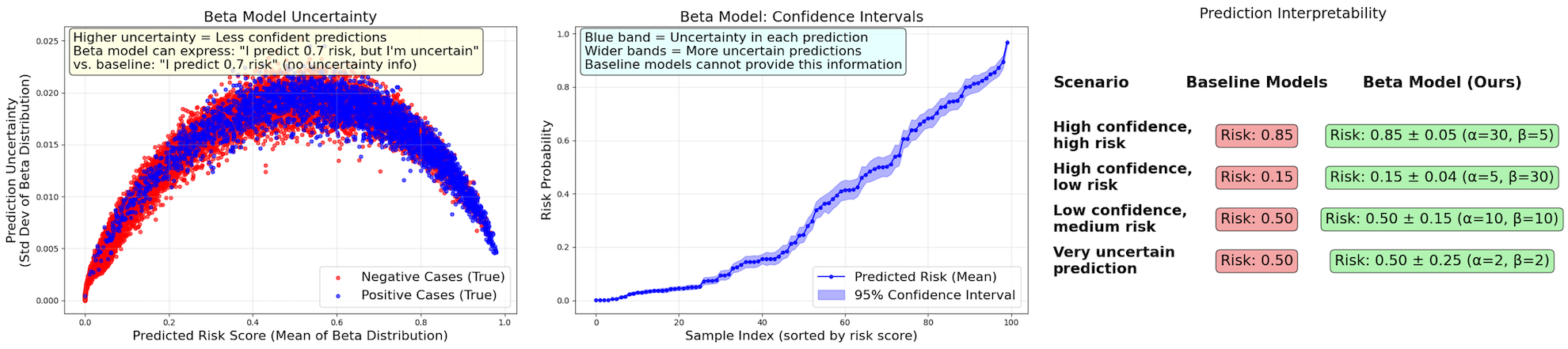}
    \caption{Uncertainty Quantification and Interpretability}
    \label{fig:interpretability_summary}
\end{figure*}

Our qualitative analysis highlights the superior interpretability and trustworthiness of our probabilistic framework. As shown in Figure~\ref{fig:interpretability_summary}, our model provides a comprehensive and practical understanding of risk that standard classifiers cannot offer. The ``Beta Model Uncertainty" plot (left) confirms the model's rational behavior, showing that prediction uncertainty is lowest for highly confident predictions and highest for ambiguous ones around a 0.5 risk score. The ``Confidence Intervals" plot (center) demonstrates that every prediction is accompanied by a 95\% confidence interval, with the interval's width directly communicating the model's certainty on a per-prediction basis. Finally, the ``Prediction Interpretability" table (right) crystallizes this key advantage, showing how our Beta model resolves the ambiguity of a baseline's ``Risk: 0.50" output by distinguishing between a low-confidence prediction (e.g., with $\alpha=10, \beta=10$) and a very uncertain one (e.g., with $\alpha=2, \beta=2$). This additional context is invaluable for any safety-critical application.

\begin{figure}[!tb]
    \centering
    \includegraphics[width=.9\linewidth]{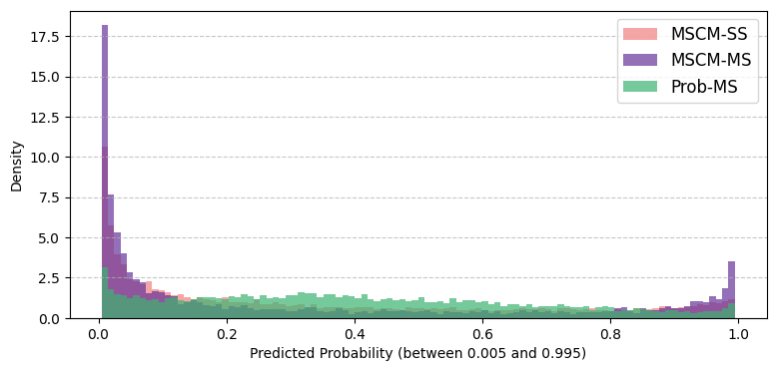}
    \caption{Analysis of Predicted Probability Distributions}
    \label{fig:probability_distributions}
\end{figure}

This nuanced, per-prediction behavior leads to a more rational distribution of predictions in aggregate (Figure~\ref{fig:probability_distributions}). While baseline models behave like overconfident black boxes with predictions heavily clustered at the extremes of 0 and 1, our model utilizes the full probability spectrum to express varying degrees of certainty. This ability to be ``less confident" is not a weakness but a hallmark of a more honest and trustworthy risk assessment tool.

\subsection{Visualizing Model Uncertainty}

\begin{figure}[!tb]
    \centering
    \begin{subfigure}[b]{0.495\textwidth}
         \centering
         \includegraphics[width=\textwidth]{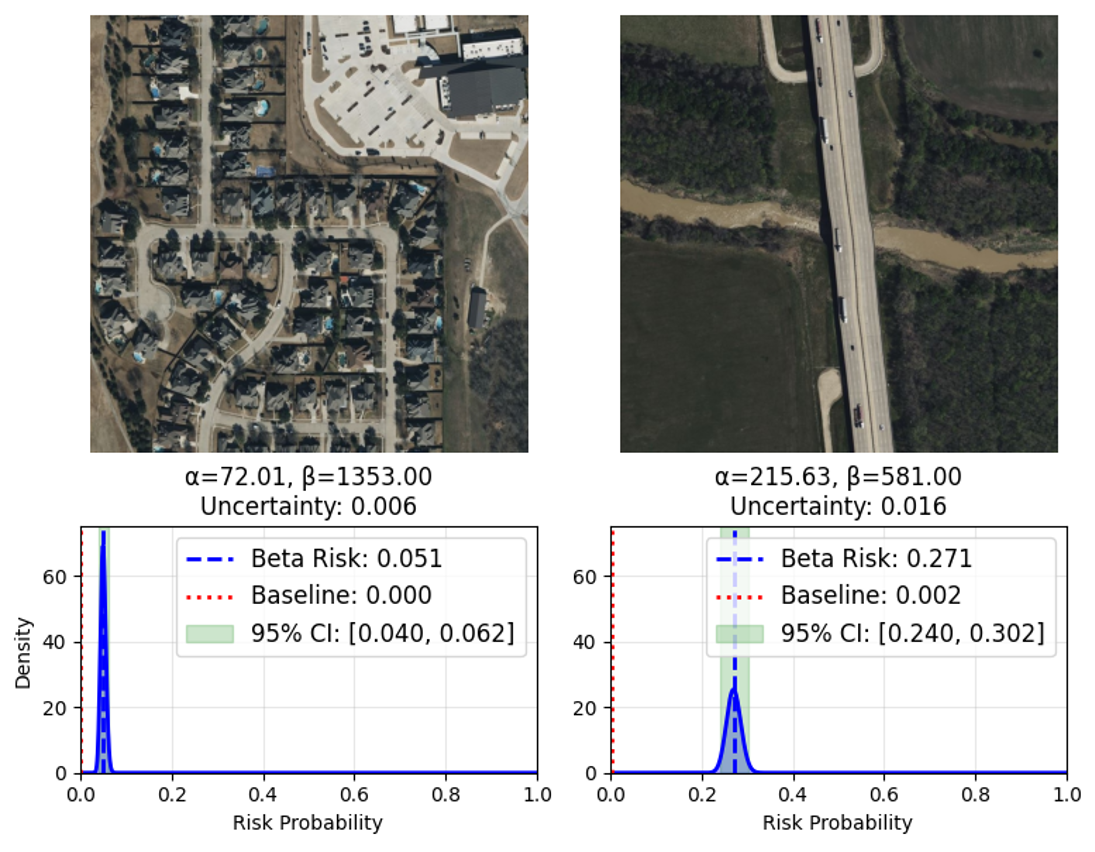}
         \caption{}
         \label{fig:tn}
     \end{subfigure}~~~
     \begin{subfigure}[b]{0.495\textwidth}
         \centering
         \includegraphics[width=\textwidth]{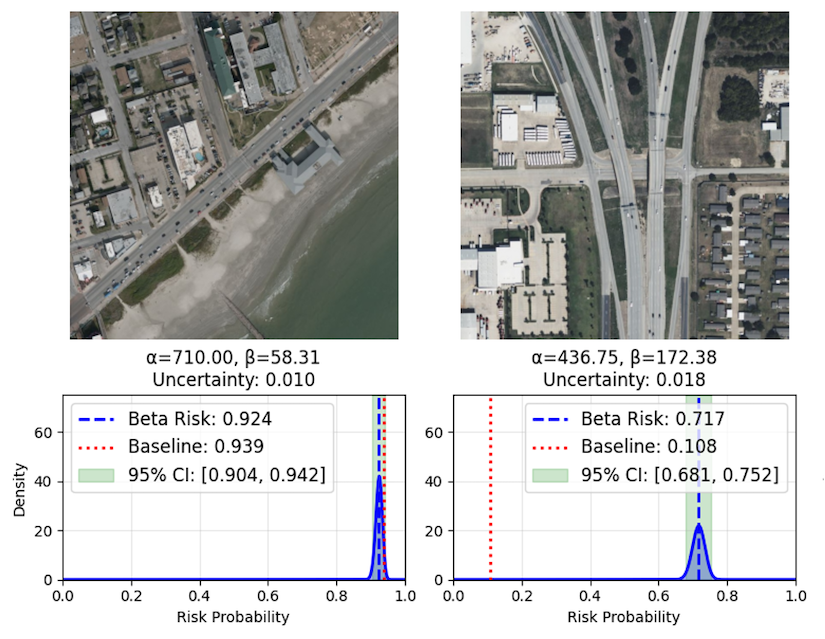}
         \caption{}
         \label{fig:tp}
     \end{subfigure}
     \caption{Qualitative Results for Unambiguous (``Easy") Cases (a: True Negatives, b: True Positives)}
     \label{fig:easy}
\end{figure}

\begin{figure}[!tb]
    \centering
    \begin{subfigure}[b]{0.495\linewidth}
         \centering
         \includegraphics[width=\linewidth]{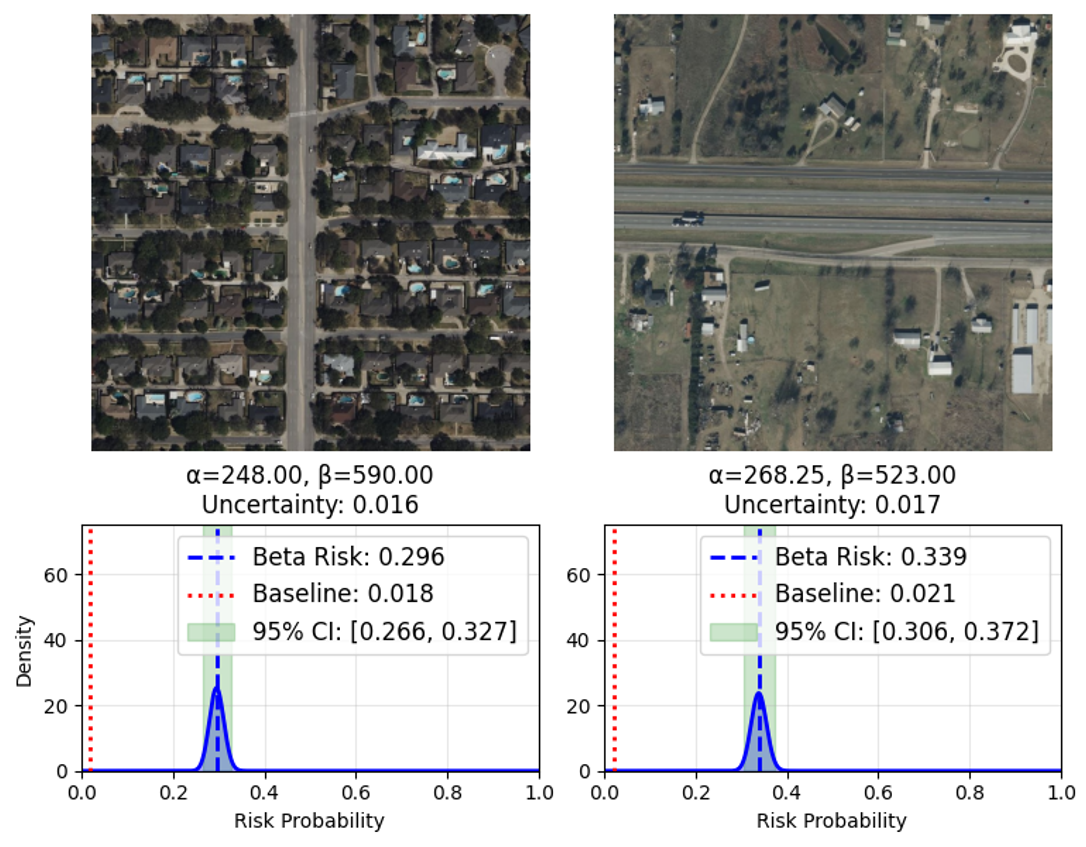}
         \caption{}
         \label{fig:fn}
     \end{subfigure}~~~
     \begin{subfigure}[b]{0.495\linewidth}
         \centering
         \includegraphics[width=\linewidth]{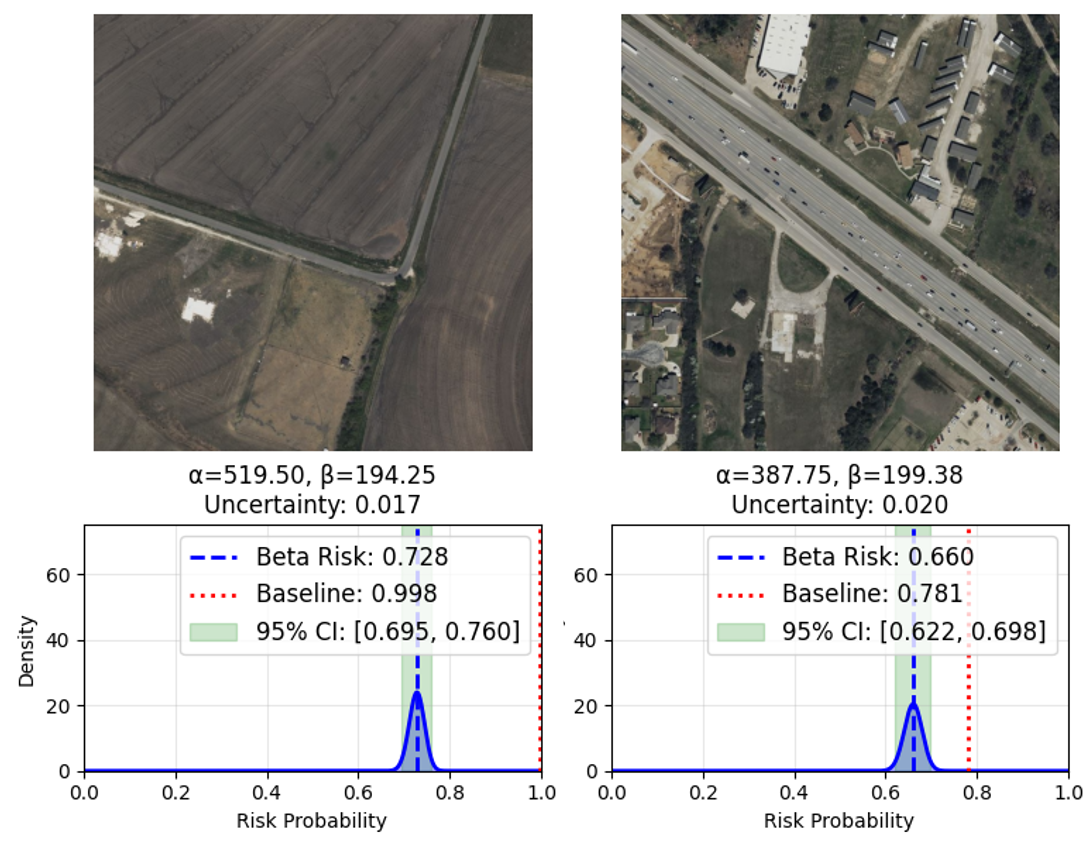}
         \caption{}
         \label{fig:fp}
     \end{subfigure}  
     \caption{Interpreting Model Behavior on Ambiguous (``Hard") Cases (a: False Negative, b: False Positive)}
     \label{fig:hard}
\end{figure}

To be a trustworthy tool for risk assessment, a model must not only make accurate predictions but also provide a reliable measure of its own uncertainty. We visualize this uncertainty using a Beta distribution for each prediction. As shown in Figure~\ref{fig:easy}, our model demonstrates well-calibrated confidence across a spectrum of cases, a crucial feature for real-world deployment.

For visually unambiguous locations, the model produces predictions with high confidence. For example, in a simple suburban neighborhood (Figure~\ref{fig:tn}, left), it predicts a low risk (0.051) with a correspondingly low uncertainty score (0.006), reflected in a sharp Beta distribution. Likewise, for a coastal road with high traffic density and high potential of distractions (Figure~\ref{fig:tp}, left), it correctly predicts a high risk (0.924) with high confidence (uncertainty of 0.010).

The model's utility is further demonstrated in more complex scenarios where it appropriately reduces its confidence. For a visually complex but safe highway overpass (Figure~\ref{fig:tn}, right), the model still correctly predicts low risk, but the wider Beta distribution indicates higher uncertainty. This nuanced confidence is also evident when assessing a complex highway interchange (Figure~\ref{fig:tp}, right); the model correctly predicts a high risk of 0.717, but acknowledges the significant uncertainty due to the challenging visual features.


Crucially, the model's rational expression of uncertainty extends to its failures, a characteristic vital for establishing trust. For false negatives (Figure \ref{fig:fn}), where the model misses a crash, the low-risk predictions are consistently paired with wider, higher-uncertainty distributions. This correctly signals that the visual evidence was ambiguous, containing conflicting features (e.g., Figure \ref{fig:fn} Left: an arterial road with many intersections within an otherwise low-risk residential area).

Similarly, for false positives (Figure \ref{fig:fp}), the model flags locations as high-risk despite no recorded crashes, but again with reduced confidence. This behavior is highly interpretable, as the model correctly identifies latent risk factors, such as sharp (L-shape) turns or high-density highways. The prediction thus reflects a successful identification of hazardous features, while the increased uncertainty correctly marks them as borderline cases. This ability to temper certainty in response to visual complexity, especially when incorrect, distinguishes our model as a more reliable and interpretable system for risk assessment.

\subsection{Case Study}
To demonstrate the practical utility of our model, we conducted a case study of the San Antonio River Walk, a major tourist destination that presents a challenging environment with a complex mix of vehicular, pedestrian, and cyclist traffic. We generated risk predictions for over 140 locations in this area using \texttt{Prob-MS} and \texttt{MSCM-MS}.

\begin{figure}[!tb]
    \centering
    \vspace{1mm}
    \includegraphics[width=0.925\linewidth]{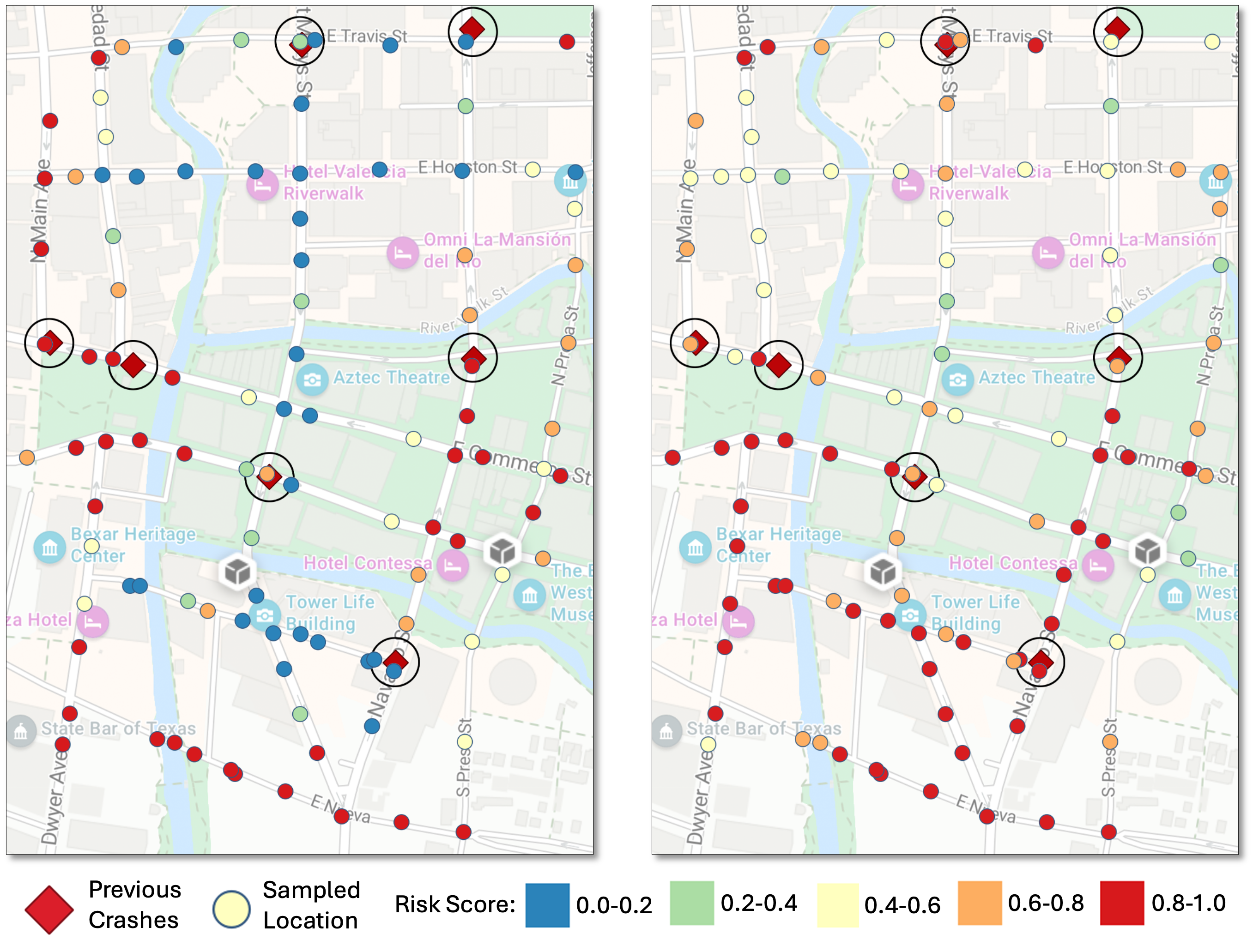}
    \caption{A case study of crash risk assessment for the San Antonio River Walk. Historical fatal crashes (red diamonds) serve as ground truth. (Left) The baseline \texttt{MSCM-MS} model exhibits low recall and spatially inconsistent predictions, with abrupt risk changes between adjacent points. (Right) Our \texttt{Prob-MS} model demonstrates superior recall by correctly identifying more crash sites and generates a more nuanced and spatially coherent risk field, providing a more realistic safety assessment. See the Result section for more.}
    \label{fig:riverwalk}
\end{figure}

The results, shown in Figure~\ref{fig:riverwalk}, highlight the superior performance of our approach. The baseline \texttt{MSCM-MS} model (middle panel) fails to identify close to half of the historical fatal crash locations (red diamonds), assigning them erroneously low risk scores. The baseline's predictions also lack spatial coherence, exhibiting sharp, unrealistic gradients between adjacent points and producing polarized risk scores with few intermediate values.

In contrast, our \texttt{Prob-MS} model (right panel) correctly assigns elevated risk scores (yellow and orange) to a greater number of the known crash sites. This is exemplified at the intersection near Navarro St and Villita St, a known fatal crash location at the bottom-right in the map. While the baseline model misses this site, ours correctly assigns the area a high-risk score.

An analysis of the satellite and ground-level imagery reveals a confluence of latent risk factors not apparent from an overhead view alone. The location, a major entry point to the River Walk, is surrounded by numerous parking facilities. Ground-level images (Figure~\ref{fig:146-Navarro-St}) show that these structures, combined with dense trees and building columns, create significant visual obstructions and blind spots for both drivers and pedestrians. This environment forces complex interactions: vehicles constantly enter and exit parking garages across wide pedestrian walkways as tourists navigate narrow sidewalks. Our model likely learned to associate this specific combination of visual clutter, unpredictable vehicle maneuvers, and high pedestrian-vehicle conflict with an elevated risk of a fatal crash.

Furthermore, our model generates a more nuanced and spatially coherent risk map where predictions transition smoothly across locations. This case study demonstrates that our model's strong quantitative performance translates into more reliable, interpretable, and actionable safety assessments for complex urban environments.

\section{Discussion and Conclusion}

Our evaluation demonstrates that the proposed probabilistic framework yields a risk assessment model that is not only more effective but also more reliable and interpretable than deterministic baselines. 
By predicting a full Beta probability distribution instead of a single point-estimate, our model learns a more nuanced and less overconfident representation of risk. This trustworthiness is reinforced by its interpretable behavior; the model’s ``mistakes" are often rational, such as flagging visually complex but historically safe highway interchanges as high-risk. This capacity to reason about visual factors and express nuanced confidence is highly valuable for practical applications, from enabling more sophisticated path planning for autonomous vehicles to allowing urban planners to confidently prioritize infrastructure improvements. Furthermore, by relying solely on publicly available satellite imagery, our method circumvents the significant privacy concerns associated with other data sources

\subsection{Ethical Considerations and Responsible Deployment} 
The ethical implications of deploying an AI tool for public safety are significant. As historical crash data may contain undiscovered biases, such as under-reporting in certain socioeconomic or geographic areas, a model used without critical oversight could perpetuate inequities. We therefore emphasize that this model is designed as a decision-support tool to augment, not replace, human expertise.

A key feature for responsible, human-in-the-loop deployment is the model's ability to signal its own uncertainty, which can serve as a bias and fairness mitigation tool. High uncertainty in any prediction (whether high-risk or low-risk) can flag regions with potential data disparities or under-reporting. For instance, a visually complex area with high uncertainty and a low-risk prediction may indicate a dangerous false negative due to a lack of historical crash data. These uncertain predictions should act as a flag for human experts to conduct a more detailed investigation, enabling a more equitable allocation of safety resources.

\subsection{Limitations}
This study has several limitations that open avenues for future research. Our model estimates static risk and does not account for dynamic variables like real-time traffic or weather; future work should focus on integrating these data streams. Our study is also geographically constrained to Texas, and validation on diverse international datasets is a critical next step to ensure generalizability. Furthermore, this work can be extended by exploring a learned weighting mechanism for the centrality and size components of our procedural labeling scheme. Finally, while our model identifies strong correlations, future work could explore methods for moving toward causal inference.

In conclusion, this work demonstrates that moving from deterministic point-estimates to a full probabilistic framework is a crucial step toward creating more reliable and trustworthy AI for public safety. By learning to predict a Beta probability distribution from satellite imagery, our model not only outperforms existing baselines in identifying high-risk locations but also provides the well-calibrated uncertainty estimates that are vital for interpretable, human-in-the-loop decision-making in applications from urban planning to autonomous navigation.
\vspace{-2.mm}
\section{Conclusion}
\vspace{-1mm}

This work presents a deep learning framework for reliable roadway risk assessment that quantifies uncertainty. Instead of a single risk score, our model predicts a full Beta probability distribution to provide a more comprehensive hazard assessment. This is achieved using a procedural labeling technique with data augmentation to supervise uncertainty, and a compound loss function that jointly optimizes for classification accuracy and probabilistic calibration.

Our model significantly outperforms existing baselines, with a 17-23\% relative improvement in recall on high-risk locations, up to 17\% in ECE on calibration, and about 11\% more stable. More importantly, it yields interpretable predictions, reliably signaling its uncertainty in ambiguous cases. By delivering a more robust and trustworthy assessment of roadway risk, our work represents a crucial step toward the responsible deployment of AI in high-stakes applications, such as public safety, urban planning, and autonomous navigation. 
\section*{Acknowledgement}

This material is partially based upon work supported by the National Science Foundation under 2401860 and 2526487. Any opinions, findings, and conclusions or recommendations expressed in this material are those of the author and the funders have no role in the study design, data collection, analysis, or preparation of this article.

\noindent Portions of this research were conducted with the advanced computing resources provided by the High Performance Computing Research Center at Texas A\&M University-San Antonio.

\bibliographystyle{IEEEtran}  
\bibliography{reference}

@article{who2023road,
  title = {Road Traffic Injuries},
  journal={World Health Organization},
  year={2023},
  author={WHO},
  url = {https://www.who.int/news-room/fact-sheets/detail/road-traffic-injuries},
  note = {Accessed: 2025-05-22}
}

@inproceedings{zulu2024enhancing,
  title={Enhancing machine learning based sql injection detection using contextualized word embedding},
  author={Zulu, Janet and Han, Bonian and Alsmadi, Izzat and Liang, Gongbo},
  booktitle={Proceedings of the 2024 ACM Southeast Conference},
  pages={211--216},
  year={2024}
}

@article{lin2022estimating,
  title={Estimating cluster masses from SDSS multiband images with transfer learning},
  author={Lin, Sheng-Chieh and Su, Yuanyuan and Liang, Gongbo and Zhang, Yuanyuan and Jacobs, Nathan and Zhang, Yu},
  journal={Monthly Notices of the Royal Astronomical Society},
  volume={512},
  number={3},
  pages={3885--3894},
  year={2022},
  publisher={Oxford University Press}
}

@inproceedings{jonnala2025exploring,
  title={Exploring the potential of large language models in public transportation: San antonio case study},
  author={Jonnala, Ramya and Liang, Gongbo and Yang, Jeong and Alsmadi, Izzat},
  Southeast={39th Annual AAAI Conference on Artificial Intelligence (AAAI) Workshop on AI for Urban Planning},
  year={2025}
}

@article{liu2022llrhnet,
  title={LLRHNet: multiple lesions segmentation using local-long range features},
  author={Liu, Liangliang and Wang, Ying and Chang, Jing and Zhang, Pei and Liang, Gongbo and Zhang, Hui},
  journal={Frontiers in Neuroinformatics},
  volume={16},
  pages={859973},
  year={2022},
  publisher={Frontiers Media SA}
}

@article{xing2023self,
  title={Self-supervised learning application on covid-19 chest x-ray image classification using masked autoencoder},
  author={Xing, Xin and Liang, Gongbo and Wang, Chris and Jacobs, Nathan and Lin, Ai-Ling},
  journal={Bioengineering},
  volume={10},
  number={8},
  pages={901},
  year={2023},
  publisher={MDPI}
}

@article{zhu2024equity,
  title={Equity in non-motorist safety: Exploring two pathways in Houston},
  author={Zhu, Chunwu and Dadashova, Bahar and Lee, Chanam and Ye, Xinyue and Brown, Charles T},
  journal={Transportation research part D: transport and environment},
  volume={132},
  pages={104239},
  year={2024},
  publisher={Elsevier}
}

@article{who2018global,
  title = {Global status report on road safety},
  journal={World Health Organization},
  year={2018},
  author={WHO},
  url = {https://www.who.int/publications/i/item/9789241565684},
  note = {Accessed: 2025-05-22}
}

@article{cisa2024critical,
  title = {Critical Infrastructure Sectors},
  author={CISA},
  journal={Cybersecurity and Infrastructure Security Agency},
  url = {https://www.cisa.gov/topics/critical-infrastructure-security-and-resilience/critical-infrastructure-sectors},
  year={2024},
  note = {Accessed: 2024-03-22}
}

@article{jaroszweski2014influence,
  title={The influence of rainfall on road accidents in urban areas: A weather radar approach},
  author={Jaroszweski, David and McNamara, Tom},
  journal={Travel behaviour and society},
  volume={1},
  number={1},
  pages={15--21},
  year={2014},
  publisher={Elsevier}
}

@article{tamerius2016precipitation,
  title={Precipitation effects on motor vehicle crashes vary by space, time, and environmental conditions},
  author={Tamerius, JD and Zhou, X and Mantilla, R and Greenfield-Huitt, T},
  journal={Weather, Climate, and Society},
  volume={8},
  number={4},
  pages={399--407},
  year={2016},
  publisher={American Meteorological Society}
}

@article{caliendo2007crash,
  author={Caliendo, Ciro and others},
  journal={Accident Analysis \& Prevention},
  volume={39},
  number={4},
  pages={657--670},
  year={2007},
  publisher={Elsevier}
}

@article{simons2014keep,
  title={Keep your eyes on the road: Young driver crash risk increases according to duration of distraction},
  author={Simons-Morton, Bruce G and Guo, Feng and Klauer, Sheila G and Ehsani, Johnathon P and Pradhan, Anuj K},
  journal={Journal of Adolescent Health},
  volume={54},
  number={5},
  pages={S61--S67},
  year={2014},
  publisher={Elsevier}
}

@article{ma2020modeling,
title = {Modeling crash risk of horizontal curves using large-scale auto-extracted roadway geometry data},
journal = {Accident Analysis \& Prevention},
volume = {144},
pages = {105669},
year = {2020},
issn = {0001-4575},
doi = {https://doi.org/10.1016/j.aap.2020.105669},
author = {Qingyu Ma and Hong Yang and Zhenyu Wang and Kun Xie and Di Yang}
}

@article{ahmed2013road,
  title={Road infrastructure and road safety},
  author={Ishtiaque Ahmed},
  journal={Transport and Communications Bulletin for Asia and the Pacific},
  volume={83},
  pages={19--25},
  year={2013}
}

@inproceedings{pembuain2019effect,
  title={The effect of road infrastructure on traffic accidents},
  author={Pembuain, Ardilson and others},
  booktitle={11th Asia Pacific Transportation and the Environment Conference (APTE 2018)},
  pages={176--182},
  year={2019},
  organization={Atlantis Press}
}

@article{huang2020highway,
title = {Highway crash detection and risk estimation using deep learning},
journal = {Accident Analysis \& Prevention},
volume = {135},
pages = {105392},
year = {2020},
issn = {0001-4575},
doi = {https://doi.org/10.1016/j.aap.2019.105392},
author = {Tingting Huang and others}
}

@article{cheng2019risk,
  title={Risk evaluation method for highway roadside accidents},
  author={Cheng, Guozhu and Cheng, Rui and Zhang, Sulu and Sun, Xiaoduan},
  journal={Advances in Mechanical Engineering},
  volume={11},
  number={1},
  pages={1687814018821743},
  year={2019},
  publisher={SAGE Publications Sage UK: London, England}
}

@article{joo2023a,
title = {A generalized driving risk assessment on high-speed highways using field theory},
journal = {Analytic Methods in Accident Research},
volume = {40},
pages = {100303},
year = {2023},
issn = {2213-6657},
doi = {https://doi.org/10.1016/j.amar.2023.100303},
author = {Yang-Jun Joo and others}
}

@INPROCEEDINGS{song2018farsa,
  author={Song, Weilian and Workman, Scott and Hadzic, Armin and Zhang, Xu and Green, Eric and Chen, Mei and Souleyrette, Reginald and Jacobs, Nathan},
  booktitle={2018 IEEE Winter Conference on Applications of Computer Vision (WACV)}, 
  title={FARSA: Fully Automated Roadway Safety Assessment}, 
  year={2018},
  volume={},
  number={},
  pages={521-529},
  doi={10.1109/WACV.2018.00063}}

@article{liang2024unveiling,
  title={Unveiling roadway hazards: Enhancing fatal crash risk estimation through multiscale satellite imagery and self-supervised cross-matching},
  author={Liang, Gongbo and Zulu, Janet and Xing, Xin and Jacobs, Nathan},
  journal={IEEE Journal of Selected Topics in Applied Earth Observations and Remote Sensing},
  volume={17},
  pages={535--546},
  year={2024},
  publisher={IEEE}
}

@article{al2012use,
  title={The use of Monte Carlo simulation in evaluating the elevator round trip time under up-peak traffic conditions and conventional group control},
  author={Al-Sharif, Lutfi and others},
  journal={Building Services Engineering Research and Technology},
  volume={33},
  number={3},
  pages={319--338},
  year={2012},
  publisher={Sage Publications Sage UK: London, England}
}

@article{jeon2016monte,
  title={Monte Carlo simulation-based traffic speed forecasting using historical big data},
  author={Jeon, Seungwoo and Hong, Bonghee},
  journal={Future generation computer systems},
  volume={65},
  pages={182--195},
  year={2016},
  publisher={Elsevier}
}

@article{de2018evaluating,
  title={Evaluating the sustainability of urban passenger transportation by Monte Carlo simulation},
  author={de Almeida Guimar{\~a}es, Vanessa and others},
  journal={Renewable and Sustainable Energy Reviews},
  volume={93},
  pages={732--752},
  year={2018},
  publisher={Elsevier}
}

@inproceedings{moosavi2019accident,
  title={Accident risk prediction based on heterogeneous sparse data: New dataset and insights},
  author={Moosavi, Sobhan and Samavatian, Mohammad Hossein and Parthasarathy, Srinivasan and Teodorescu, Radu and Ramnath, Rajiv},
  booktitle={Proceedings of the 27th ACM SIGSPATIAL International Conference on Advances in Geographic Information Systems},
  pages={33--42},
  year={2019}
}

@inproceedings{he2021inferring,
  title={Inferring high-resolution traffic accident risk maps based on satellite imagery and GPS trajectories},
  author={He, Songtao and Sadeghi, Mohammad Amin and Chawla, Sanjay and Alizadeh, Mohammad and Balakrishnan, Hari and Madden, Samuel},
  booktitle={Proceedings of the IEEE/CVF International Conference on Computer Vision},
  pages={11977--11985},
  year={2021}
}

@inproceedings{najjar2017combining,
  title={Combining satellite imagery and open data to map road safety},
  author={Najjar, Alameen and others},
  booktitle={Proceedings of the AAAI Conference on Artificial Intelligence},
  volume={31},
  number={1},
  year={2017}
}

@article{krizhevsky2012imagenet,
  title={Imagenet classification with deep convolutional neural networks},
  author={Krizhevsky, Alex and others},
  journal={Advances in neural information processing systems},
  volume={25},
  year={2012}
}

@inproceedings{chen2023ai,
  title={Is this ai trained on credible data? the effects of labeling quality and performance bias on user trust},
  author={Chen, Cheng and Sundar, S Shyam},
  booktitle={Proceedings of the 2023 CHI Conference on Human Factors in Computing Systems},
  pages={1--11},
  year={2023}
}

@article{li2024label,
  title={Label Bias: A Pervasive and Invisibilized Problem},
  author={Li, Yunyi and others},
  journal={Notices of the American Mathematical Society},
  volume={71},
  number={8},
  pages={1069--1077},
  year={2024}
}

@inproceedings{guo2017calibration,
  title={On calibration of modern neural networks},
  author={Guo, Chuan and Pleiss, Geoff and Sun, Yu and Weinberger, Kilian Q},
  booktitle={International conference on machine learning},
  pages={1321--1330},
  year={2017},
  organization={PMLR}
}

@inproceedings{liang2020imporved,  
  title={Improved Trainable Calibration Method for Neural Networks on Medical Imaging Classification},  
  author={Liang, Gongbo and Zhang, Yu and Wang, Xiaoqin and Jacobs, Nathan},  
  booktitle={British Machine Vision Conference (BMVC)},  
  year={2020}
}

@article{hinton2015distilling,
  title={Distilling the Knowledge in a Neural Network},
  author={Geoffrey E. Hinton and others},
  journal={ArXiv},
  year={2015},
  volume={abs/1503.02531},
}

@inproceedings{kumar2018trainable,
  title={Trainable calibration measures for neural networks from kernel mean embeddings},
  author={Kumar, Aviral and others},
  booktitle={International Conference on Machine Learning},
  pages={2805--2814},
  year={2018}
}

@inproceedings{pereyra2017regularizing,
  title={Regularizing neural networks by penalizing confident output distributions},
  author={Pereyra, Gabriel and Tucker, George and Chorowski, Jan and Kaiser, {\L}ukasz and Hinton, Geoffrey},
  booktitle={International Conference on Learning Representations},
  year={2017}
}

@inproceedings{chidambaram2023on,
  title={On the Limitations of Temperature Scaling for Distributions with Overlaps},
  author={Muthuraman Chidambaram and Rong Ge},
  booktitle={International Conference on Learning Representations},
  year={2023},
}

@inproceedings{he2016deep,
  title={Deep residual learning for image recognition},
  author={He, Kaiming and Zhang, Xiangyu and Ren, Shaoqing and Sun, Jian},
  booktitle={Proceedings of the IEEE conference on computer vision and pattern recognition},
  pages={770--778},
  year={2016}
}

@article{loshchilov2017decoupled,
  title={Decoupled weight decay regularization},
  author={Loshchilov, Ilya and Hutter, Frank},
  journal={arXiv preprint arXiv:1711.05101},
  year={2017}
}

@article{loshchilov2016sgdr,
  title={Sgdr: Stochastic gradient descent with warm restarts},
  author={Loshchilov, Ilya and Hutter, Frank},
  journal={arXiv preprint arXiv:1608.03983},
  year={2016}
}

@article{cramer1928on,
    author = {Harald Cramér},
    title = {On the composition of elementary errors},
    journal = {Scandinavian Actuarial Journal},
    volume = {1928},
    number = {1},
    pages = {13--74},
    year = {1928},
    publisher = {Taylor \& Francis}
}

@article{csiszar1975divergence,
  title={I-divergence geometry of probability distributions and minimization problems},
  author={Csisz{\'a}r, Imre},
  journal={The annals of probability},
  pages={146--158},
  year={1975},
  publisher={JSTOR}
}

@article{vaserstein1969markov,
  title={Markov processes over denumerable products of spaces, describing large systems of automata},
  author={Vaserstein, Leonid Nisonovich},
  journal={Problemy Peredachi Informatsii},
  volume={5},
  number={3},
  pages={64--72},
  year={1969},
  publisher={Russian Academy of Sciences, Branch of Informatics, Computer Equipment and~…}
}

@article{gu2022multivariate,
  title={Multivariate analysis of roadway multi-fatality crashes using association rules mining and rules graph structures: A case study in China},
  author={Gu, Chenwei and Xu, Jinliang and Gao, Chao and Mu, Minghao and E, Guangxun and Ma, Yongji},
  journal={Plos one},
  volume={17},
  number={10},
  pages={e0276817},
  year={2022},
  publisher={Public Library of Science San Francisco, CA USA}
}

@article{carrodano2024data,
  title={Data-driven risk analysis of nonlinear factor interactions in road safety using Bayesian networks},
  author={Carrodano, Cinzia},
  journal={Scientific Reports},
  volume={14},
  number={1},
  pages={18948},
  year={2024},
  publisher={Nature Publishing Group UK London}
}

\clearpage 

\section*{Supplementary Materials}

\appendix
\setcounter{figure}{0}
\setcounter{table}{0}
\renewcommand{\thetable}{S\arabic{table}}
\renewcommand{\thefigure}{S\arabic{figure}}

\subsection*{San Antonio River Walk Ground-Level Imagery}
\begin{figure*}[!h]
    \centering
    \begin{subfigure}[b]{0.41\textwidth}
         \centering
         \includegraphics[width=\textwidth]{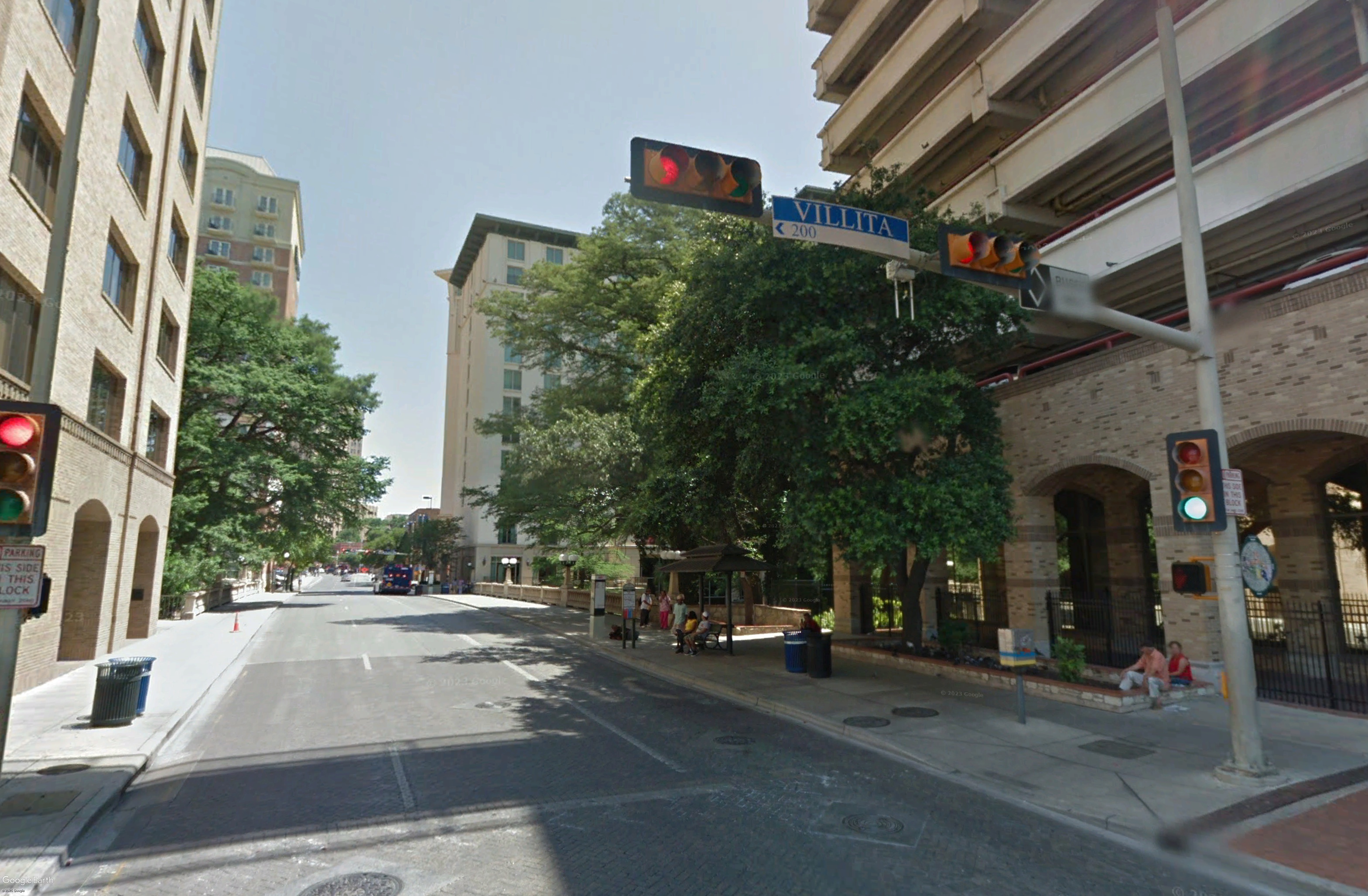}
         \caption{North}
         \label{fig:146-Navarro-St_N}
         \vspace{3mm}
     \end{subfigure}~~~~~~~
     \begin{subfigure}[b]{0.41\textwidth}
         \centering
         \includegraphics[width=\textwidth]{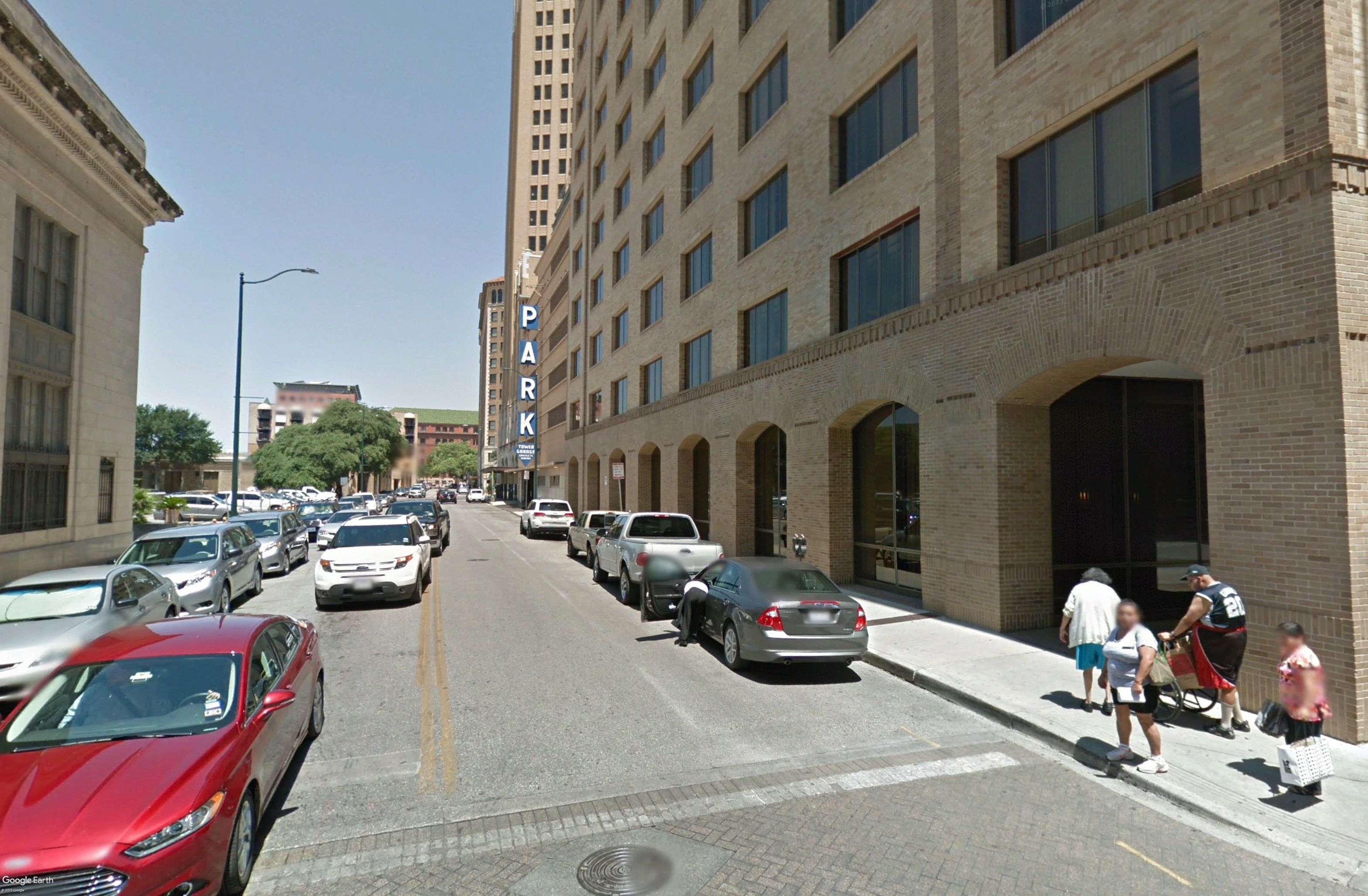}
         \caption{West}
         \label{fig:146-Navarro-St_W}
         \vspace{3mm}
     \end{subfigure}     
     
     \begin{subfigure}[b]{0.41\textwidth}
         \centering
         \includegraphics[width=\textwidth]{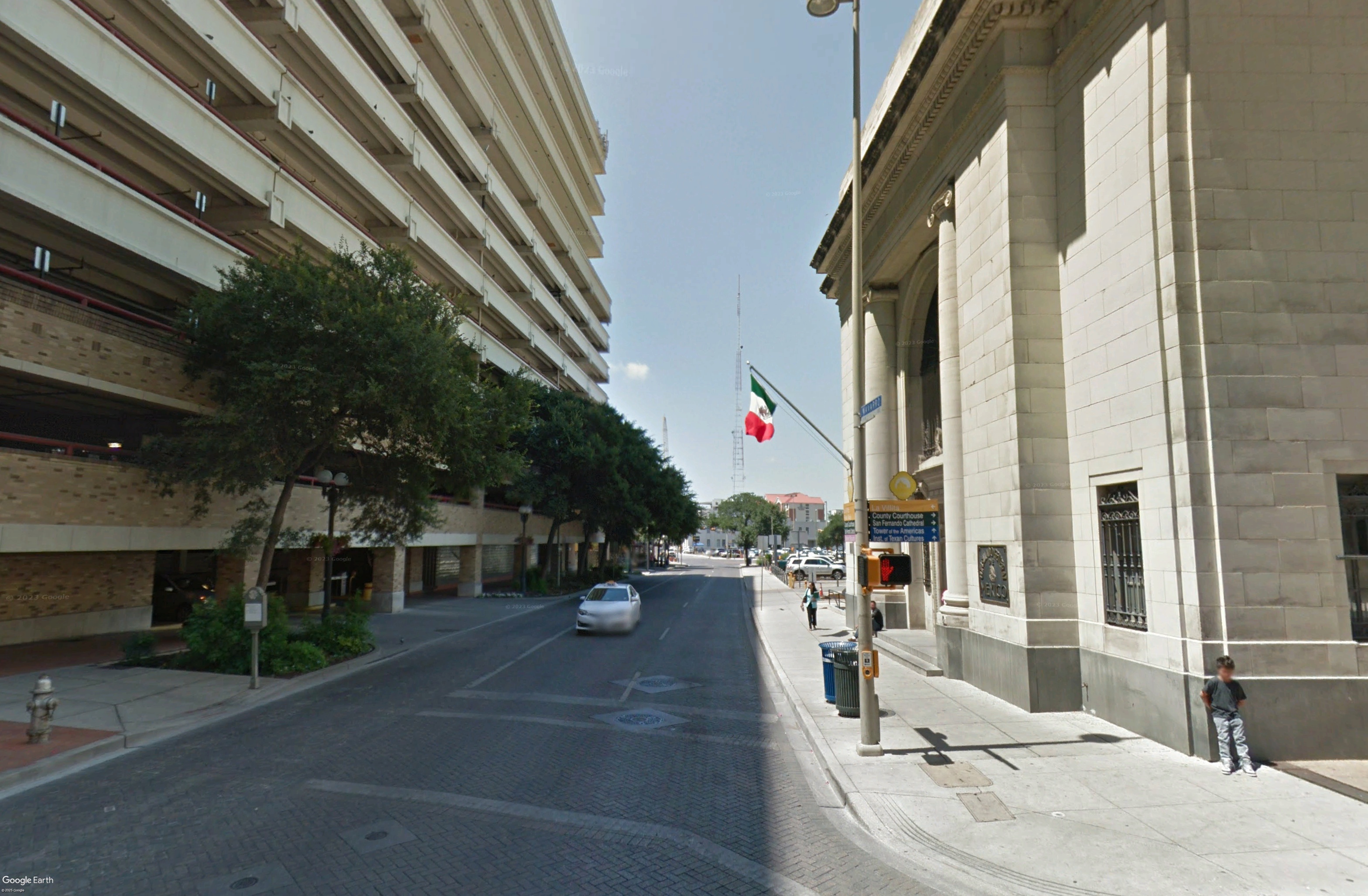}
         \caption{South}
         \label{fig:146-Navarro-St_S}
     \end{subfigure}~~~~~~~
     \begin{subfigure}[b]{0.41\textwidth}
         \centering
         \includegraphics[width=\textwidth]{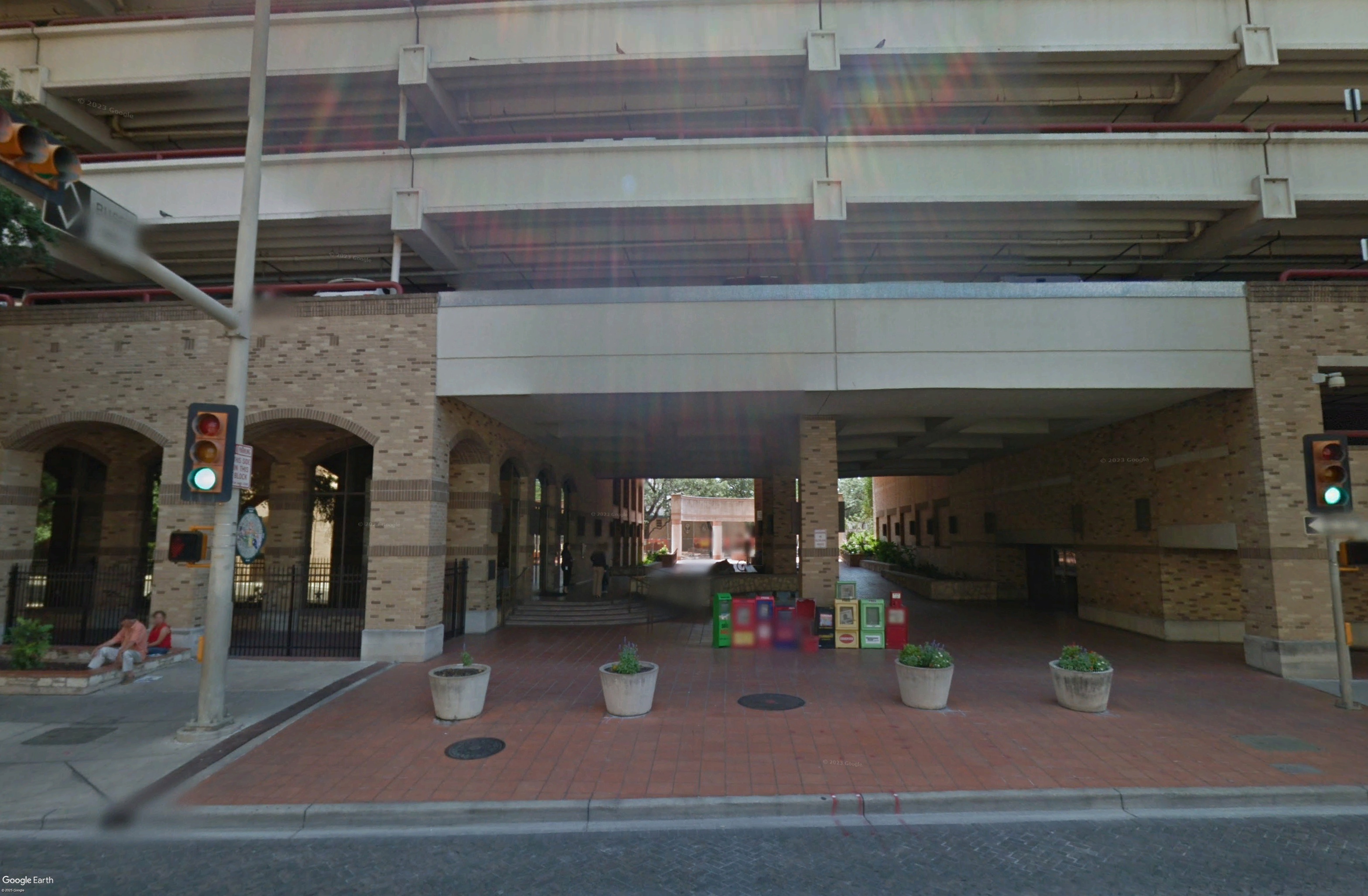}
         \caption{East}
         \label{fig:146-Navarro-St_E}
     \end{subfigure}
     \caption{Ground-level images for the four directions at 146-Navarro-St, San Antonio, TX, USA, one main entry point to the San Antonio River Walk area.}
     \label{fig:146-Navarro-St}
\end{figure*}

Figure~\ref{fig:146-Navarro-St} shows the ground-level image at 146-Navarro-St, San Antonio, TX, USA, one main entry point to the San Antonio River Walk area. A previous fatal crash also occurred at this location (i.e., the bottom-right one in Figure~\ref{fig:riverwalk}). 

\subsection*{Dataset}
This study utilizes the comprehensive, multi-scale satellite imagery dataset provided by MSCM~\cite{liang2024unveiling}. The dataset covers diverse regions in Texas, USA, including the Gulf Coast, Hill Country, and Prairies and Lakes regions. 

\begin{figure*}[!tb]
    \centering    \includegraphics[width=0.875\textwidth]{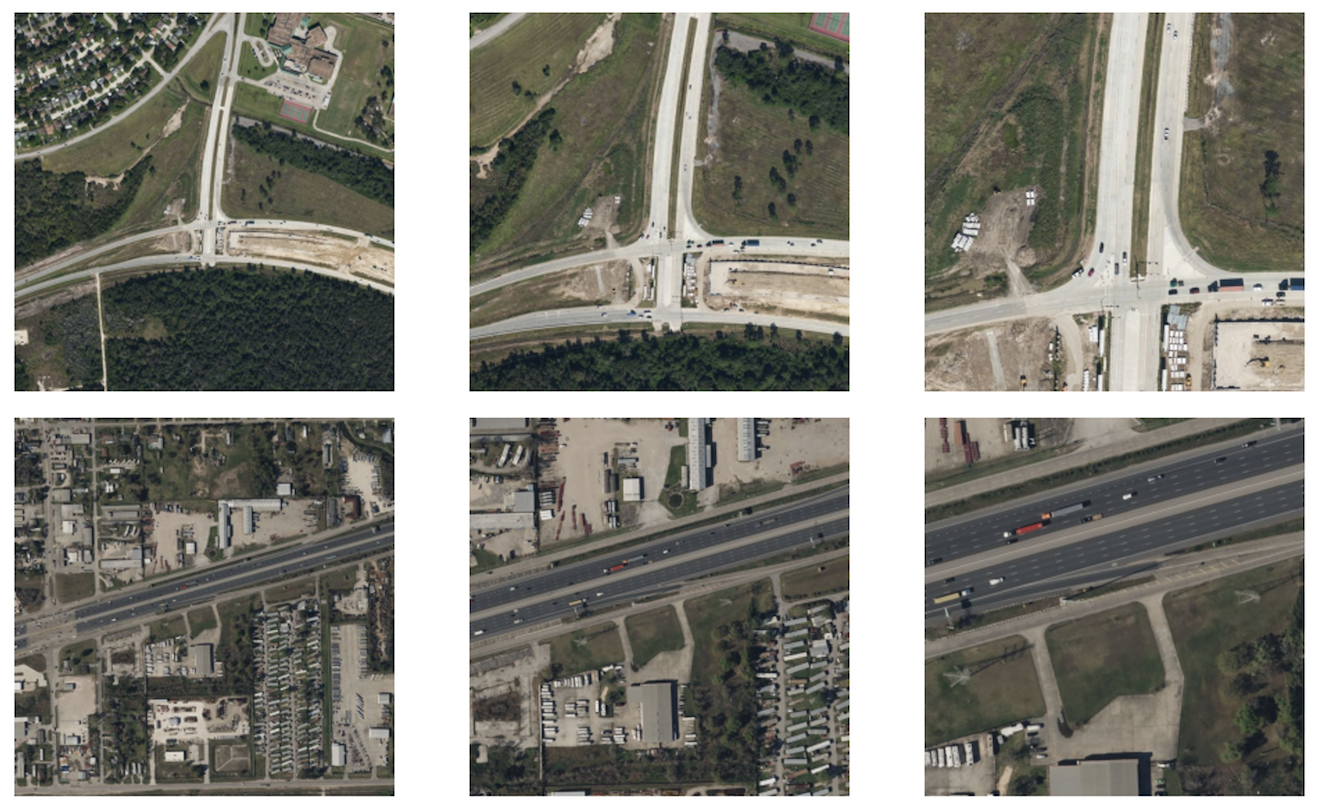}   
    \caption{Multi-Scale Satellite Imagery Inputs with Three Detail Levels. From left to right: 1.1943 m/pixel, 0.5972 m/pixel, and 0.2986 m/pixel.}
    \label{fig:dataset}
\end{figure*}

The dataset contains a total of 240,828 satellite images. The images for each location are provided at three distinct levels of detail, each with a resolution of $768\times 768$ pixels. The images for each location are provided at three distinct levels of detail: 1.1943 m/pixel, 0.5972 m/pixel, and 0.2986 m/pixel. 
Examples of this multi-scale imagery are shown in Figure~\ref{fig:dataset}.

The data is sampled from 80,276 distinct locations, which are categorized into positive and negative classes. The \textit{positive} class consists of 16,451 locations where at least one fatal crash occurred between 2010 and 2020; of these, 1,185 locations experienced multiple fatal crashes within a 50-meter radius. The remaining locations serve as the \textit{negative} class, having no recorded fatal crashes through the end of 2020. These negative samples were selected using specific criteria, ensuring they were within 1250 meters of a fatal crash location but at least 250 meters away from any such site. To create a challenging learning environment, approximately 70\% of the negative samples were designated as hard negatives by sampling them along primary and secondary roads. The other 30\% were sampled randomly to represent a broader variety of environments, including open spaces.

\subsection*{Wasserstein-2 Surrogate Analysis}
\begin{figure*}[!tb]
    \centering
    \begin{subfigure}[b]{0.423\textwidth}
         \centering
         \includegraphics[width=\textwidth]{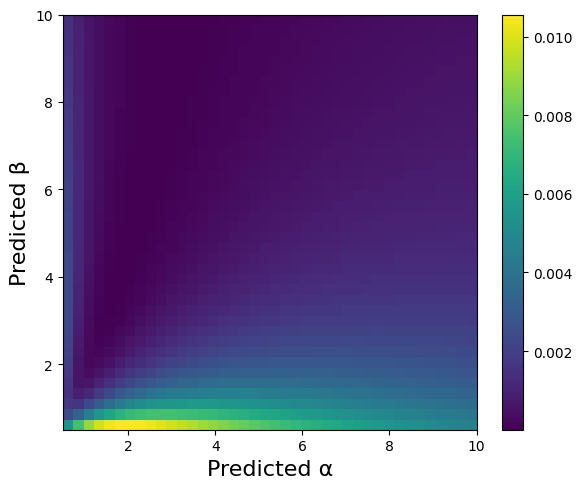}
         \label{fig:abs_error}
     \end{subfigure}~~~~~~~~~~~~~~~~
     \begin{subfigure}[b]{0.43\textwidth}
         \centering
         \includegraphics[width=\textwidth]{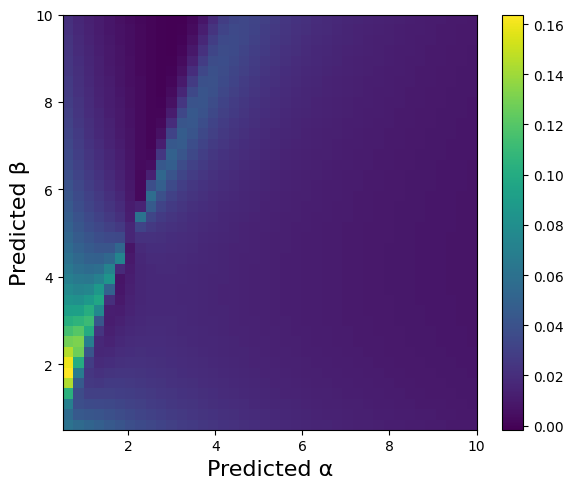}
         \label{fig:rel_error}
     \end{subfigure}     
     \caption{Comparison between true and surrogate Wasserstein-2 distance for Beta distributions. \textbf{Left:} Absolute Error. \textbf{Right:} Relative Error.}
     \label{fig:w2_error}
\end{figure*}

To evaluate the accuracy of our surrogate Wasserstein-2 loss, we computed the true squared Wasserstein-2 distance between a fixed target distribution Beta(2, 5) and a range of predicted Beta distributions with $\alpha, \beta \in [0.5, 10]$. For each pair of predicted parameters, the true distance was estimated via numerical integration of the quantile functions, while the surrogate distance was computed using the closed-form mean–variance expression defined in our loss. The resulting absolute and relative differences are visualized in Figure~\ref{fig:w2_error}. Both plots demonstrate that the surrogate closely matches the true distance, with errors typically on the order of $10^{-3}$ to $10^{-2}$ and only slightly increasing in extreme parameter regions.

\subsection*{Implementation Details}
All models are built upon a ResNet-50 backbone and trained for 75 epochs using the AdamW~\cite{loshchilov2017decoupled} optimizer with a CosineAnnealingWarmRestarts~\cite{loshchilov2016sgdr} learning rate schedule. The initial learning rate for the backbone and classifier was set to $1e^{-4}$, while the distribution learning head used a rate of 0.02. The batch size was 128 for single-scale models and 48 for multi-scale models. Based on our hyperparameter analysis, the final weights for our compound loss function were set to $\lambda_1=5$ and $\lambda_2=1$ to prioritize recall for this safety-critical task. The same data augmentation pipeline are used for all models. The random seeds are set to 0. All experiments were conducted on two NVIDIA A100 GPUs. The rest of this section, provided the complete list of hyperparameters and data augmentation settings. 

\subsubsection*{Model Architecture}
\begin{itemize}
    \item \textbf{Backbone:} All models use a ResNet-50 architecture with weights pre-trained on ImageNet.
    \item \textbf{Modification:} A 1x1 convolutional layer was inserted before the final Global Average Pooling .
\end{itemize}

\subsubsection*{MSCM Pre-training and Pre-trained Checkpoint Selection}
The MSCM weights, used to initialize our proposed model and the corresponding baselines, were generated by following the pre-training procedure described in the original work.
\begin{itemize}
    \item \textbf{Setup:} The pre-training used a contrastive learning approach with an InfoNCE and classification loss. It was run for 25 epochs with a batch size of 64, using the AdamW optimizer with a learning rate of $10^{-3}$ and a CosineAnnealingWarmRestarts schedule ({T\_0=10}, {T\_mult=2}).
    \item \textbf{Checkpoint Selection:} A model checkpoint was saved after each pre-training epoch. To select the optimal checkpoint, each of the 25 checkpoints was used to initialize a single-scale classification model, which was then fine-tuned for 5 epochs on the downstream task (batch size 128, AdamW, learning rate $10^{-4}$, {CosineAnnealingWarmRestarts} schedule). The checkpoint that yielded the best performance after this short fine-tuning process was selected for all main experiments.
\end{itemize}

\subsubsection*{Main Training for Proposed Probabilistic Models}
\begin{itemize}
    \item \textbf{Optimizer:} AdamW.
    \item \textbf{Learning Rates (LR):} We used different learning rates for distinct parts of the model:
    \begin{itemize}
        \item Feature Extractor Backbone: $10^{-4}$
        \item Beta Distribution Learning Head: $0.02$
        \item Auxiliary Classification Head: $10^{-4}$
    \end{itemize}
    \item \textbf{LR Schedule:} {CosineAnnealingWarmRestarts} with scheduler parameters \texttt{T\_0=10} and \texttt{T\_mult=2}.
    \item \textbf{Epochs:} All models were trained for 75 epochs. The epoch with the highest accuracy on the test set is chosen as the best model.
    \item \textbf{Batch Size:} 128 for single-scale models; 48 for multi-scale models.
\end{itemize}

\subsubsection*{Loss Function and Hyperparameters}
\begin{itemize}
    \item \textbf{Compound Loss Weights:} The reported results use $\lambda_1=5$ (for BCE loss) and $\lambda_2=1$ (for Wasserstein loss).
    \item \textbf{BCE Loss:} Class imbalance was handled using inverse frequency weights applied to the BCE loss: [1.25948, 4.85382].
    \item \textbf{Procedural Beta Distribution Parameters:}
    \begin{itemize}
        \item base\_K: 22.0
        \item $\epsilon$: 0.08
        \item min\_positive\_risk\_mean: 0.18
        \item min\_concentration\_positives: 18.0
        \item Influence Score Weights: \texttt{weight\_distance=0.7}, \texttt{weight\_crop\_size=0.3}.
    \end{itemize}
\end{itemize}

\subsubsection*{Data Augmentation}
The augmentation pipelines were used for our proposed models versus the baselines.
\begin{itemize}
    \item \textbf{For Proposed Probabilistic Models:}
    \begin{itemize}
        \item Random Crop Ratio: (0.5, 1.0) 
        \item Random horizontal flip (p=0.5)
        \item Random vertical flip (p=0.5)
        \item Random rotations (from -90 to 90 degrees)
        \item ColorJitter (brightness/contrast/saturation: [0.6, 1.4], hue: [0.0, 0.1])
    \end{itemize}
    \item \textbf{For Baseline Models:}
    \begin{itemize}
        \item Random Crop Ratio: (0.3, 1.0) 
        \item Random horizontal flip (p=0.5)
        \item Random vertical flip (p=0.5)
        \item Random rotations (from -90 to 90 degrees)
        \item ColorJitter (brightness/contrast/saturation: [0.6, 1.4], hue: [0.0, 0.1])
    \end{itemize}
\end{itemize}

\subsubsection*{Hardware}
\begin{itemize}
    \item All experiments were conducted on two NVIDIA A100 GPUs, each with 40GB of memory.
\end{itemize}

\begin{figure*}[!tb]
    \centering
    \includegraphics[width=.975\linewidth]{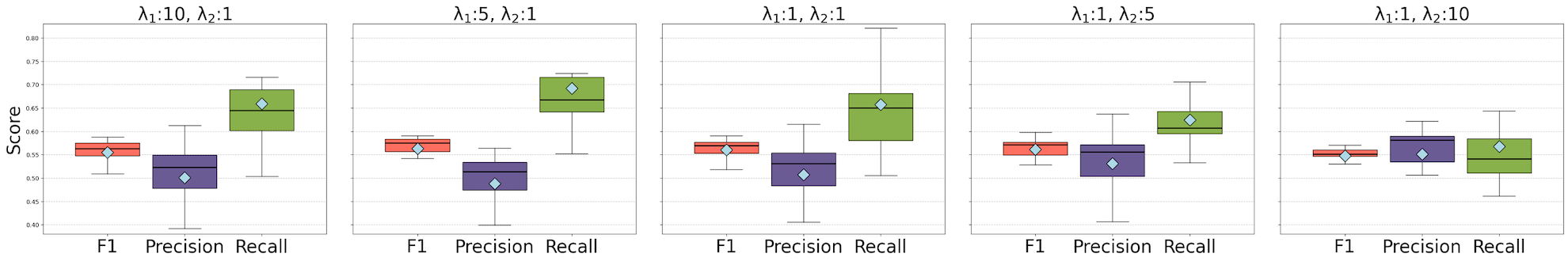}
    \caption{Hyperparameter Analysis: Precision-Recall Trade-off}
    \label{fig:weight_effects}
\end{figure*}

\subsection*{Hyperparameter Analysis: Effect of Loss Weights}

To understand how the components of our compound loss function influence model behavior, we conducted a hyperparameter analysis on the loss weights, $\lambda_1$ (for the classification loss, $\mathcal{L}_{BCE}$) 
and $\lambda_2$ (for the Beta distribution loss, $\mathcal{L}_{W_2^2}$). 
Table~\ref{table:result-weight} demonstrates that these weights serve as a practical lever to tune the model's predictive trade-offs for different application needs. See the supplementary materials for a detailed analysis.

\begin{table}[!tb]
    \footnotesize 
	\centering
	\caption{Ablation Study on Loss Weights}
    
    \begin{tabular}{c|c||c|c|c} \hline \hline
    \textbf{$\lambda_1$} & \textbf{$\lambda_2$} &\textbf{F1 Score ($\uparrow$)} & \textbf{Precision ($\uparrow$)} & \textbf{Recall ($\uparrow$)}  \\\hline\hline
    
    \textbf{10} & \textbf{1} & $0.5880$ & $0.5358$ & $0.6514$  \\\hline
    \textbf{5} & \textbf{1} & $0.5981$ & $0.5607$ & $0.6409$  \\\hline
    \textbf{1} & \textbf{1} & $0.5905$ & $0.5406$ & $0.6505$  \\\hline        
    \textbf{1} & \textbf{5} & $0.5979$ & $0.5932$ & $0.6026$  \\\hline
    \textbf{1} & \textbf{10} & $0.5855$ & $0.5577$ & $0.6163$  \\\hline\hline
        
	\end{tabular}
	\label{table:result-weight}
\end{table}

Our analysis confirms two key trends. First, increasing the weight of the classification loss ($\lambda_1$) makes the model prioritize \textbf{Recall}. As shown in Table~\ref{table:result-weight}, increasing $\lambda_1$ to 10 yields the highest recall score of 0.6514. This indicates that a stronger emphasis on the classification task pushes the model to more aggressively identify all potential high-risk locations, which is critical for safety applications.

Conversely, increasing the weight of the Beta distribution loss ($\lambda_2$) encourages a more \textbf{balanced and precise} model. The box plots in Figure~\ref{fig:weight_effects}, which show the performance distribution over 25 epochs, provide additional insight. They visually confirm that as $\lambda_2$ increases, the median precision rises while recall moderately decreases, bringing the two metrics into closer alignment. This is exemplified by the $\lambda_1=1, \lambda_2=5$ configuration, which achieves the highest precision of all tested settings (0.5932) while maintaining a strong recall (0.6026), as detailed in Table~\ref{table:result-weight}. This demonstrates that a stronger emphasis on the distribution-matching loss component encourages a more conservative model that makes fewer, but more accurate, high-risk predictions.

This analysis provides clear guidance for hyperparameter selection based on the desired outcome. For a safety-critical system where failing to identify a hazard is the worst-case scenario, a higher $\lambda_1$ is optimal. For applications requiring high confidence in positive predictions to efficiently allocate resources, a higher $\lambda_2$ would be chosen. For the main results reported in this paper, we selected the $\lambda_1=5, \lambda_2=1$ configuration, as it achieved the highest F1-score and maintained a strong recall, offering an excellent balance for our primary task.

\end{document}